\documentclass[journal]{IEEEtran}
\usepackage{amsmath,amsfonts}
\usepackage{algorithm}
\usepackage{algpseudocode}
\usepackage{hyperref}
\usepackage{array}
\usepackage[caption=false,font=normalsize,labelfont=sf,textfont=sf]{subfig}
\usepackage{textcomp}
\usepackage{stfloats}
\usepackage{url}
\usepackage{verbatim}
\usepackage{graphicx}
\usepackage{cite}
\usepackage{color}
\usepackage{multirow}
\usepackage[normalem]{ulem}
\useunder{\uline}{\ul}{}
\def\ie{{\em i.e.}}
\def\eg{{\em e.g.}}

\newcommand{\figref}[1]{Fig. \ref{#1}}
\newcommand{\tabref}[1]{Tab. \ref{#1}}

\newcommand{\bs}[1]{\boldsymbol{\texttt{#1}}}
\usepackage{balance}

\begin{document}

\title{Sensitivity Decouple Learning for Image Compression Artifacts Reduction}

\author{Li~Ma,
        Yifan~Zhao,
        Peixi~Peng,
        and~Yonghong~Tian,~\IEEEmembership{Fellow, IEEE}
\IEEEcompsocitemizethanks{
\IEEEcompsocthanksitem Li Ma and Yifan Zhao contributed equally. Corresponding authors: Peixi Peng and Yonghong Tian.
\IEEEcompsocthanksitem Li Ma is with the Huawei Technologies Company, Ltd, Shenzhen, 518129, China, (e-mail: mali93@huawei.com).
\IEEEcompsocthanksitem Yifan Zhao is with the State Key Laboratory of Virtual Reality Technology and Systems, School of Computer Science and Engineering, Beihang University, Beijing, 100191, China. (e-mail: zhaoyf@buaa.edu.cn)
\IEEEcompsocthanksitem Peixi Peng, and Yonghong Tian are with the School of Electronics and Computer Engineering and the School of Computer Science, Peking University, Beijing, 100871, China, and the PengCheng Laboratory, Shenzhen, 518055, China, (e-mail: pxpeng@pku.edu.cn; yhtian@pku.edu.cn).
}}

\maketitle

\begin{abstract}
With the benefit of deep learning techniques, recent researches have made significant progress in image compression artifacts reduction. Despite their improved performances, prevailing methods only focus on learning a mapping from the compressed image to the original one but ignore the intrinsic attributes of the given compressed images, which greatly harms the performance of downstream parsing tasks. Different from these methods, we propose to decouple the intrinsic attributes into two complementary features for artifacts reduction,~\ie, the compression-insensitive features to regularize the high-level semantic representations during training and the compression-sensitive features to be aware of the compression degree. To achieve this, we first employ adversarial training to regularize the compressed and original encoded features for retaining high-level semantics, and we then develop the compression quality-aware feature encoder for compression-sensitive features. Based on these dual complementary features, we propose a Dual Awareness Guidance Network (DAGN) to utilize these awareness features as transformation guidance during the decoding phase. In our proposed DAGN, we develop a cross-feature fusion module to maintain the consistency of compression-insensitive features by fusing compression-insensitive features into the artifacts reduction baseline. 
{
Our method achieves an average 2.06 dB PSNR gains on BSD500, outperforming state-of-the-art methods, and only requires 29.7 ms to process one image on BSD500. Besides, the experimental results on LIVE1 and LIU4K also demonstrate the efficiency, effectiveness, and superiority of the proposed method in terms of quantitative metrics, visual quality, and downstream machine vision tasks.
}
\end{abstract}


\section{Introduction}
\IEEEPARstart{T}{he} explosive growth of multimedia data, especially images and videos, brings significant challenges to bandwidth and resources. Thus, lossy compression methods have become widely applied techniques to reduce information redundancy and save storage space and transmission time.  However, due to the inevitable information loss in these compression operations, complex artifacts, such as blocking and ringing artifacts, are usually introduced into compressed images and videos. These artifacts reduce the visual quality and the performance of downstream parsing tasks, especially at high compression ratios.

\begin{figure}
    \centering
    \includegraphics[width=1.00\linewidth]{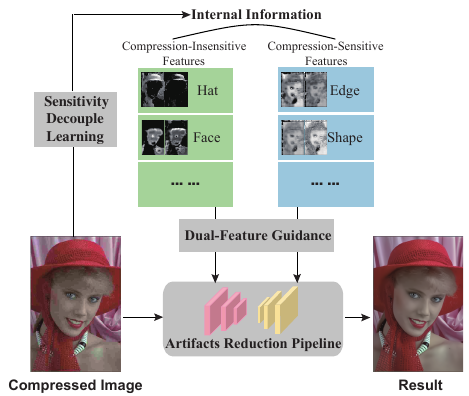}
    \caption{The motivation of the proposed sensitivity decouple learning, which explicitly mines the intrinsic attributes from the given compressed image. The proposed method decouples the intrinsic image attributes into compression-insensitive features for high-level semantics (\eg, hat and face) and compression-sensitive features for low-level cues (\eg, edge and shape).}
    \label{fig:motivation}
\end{figure}
Compression artifacts reduction tasks aim at reducing the artifacts caused by lossy compression, \emph{i.e.}, predicting original (uncompressed) images or videos from compressed ones. 
Prevailing compression artifacts reduction methods could fall into three classes: filter-based methods \cite{zhai2008efficient,yoo2014post,foi2007pointwise}, prior-based methods \cite{zhang2016concolor,liu2016data,bredies2012total,chang2013reducing,song2020compressed,zhang2016low,zhang2013compression,liu2018graph}, and learning-based methods \cite{zhang2022deep,jiang2021towards,vandal2018generating,RNANzhang2019residual,dncnnzhang2017beyond,li2017efficient,wang2019cfsnet,kim2020agarnet,chen2020adaptive,jin2020dual,qiu2020deep,zhang2020residual,fu2019jpeg,wang2016d3,guo2016building,chen2018dpw,zhang2018dmcnn,sun2020reduction,galteri2017deep,guo2017one,galteri2019deep,chen2018deep,zheng2019implicit,li2020learning,kim2019pseudo,ma2021reducing,ma2019residual,wang2017novel,liu2020comprehensive,ARCNNdong2015compression, mu2020graph,chen2021feature,zha2021triply,chen2021pre,liang2021swinir,cao2021video,wang2021uformer,zhang2019deep,dong2018denoising,jo2021rethinking,zhang2021plug,fu2021learning,fu2021model,liu2021underexposed,zhang2017learning,liu2019knowledge}. 
The pioneer works \cite{zhai2008efficient,yoo2014post,foi2007pointwise} design handcrafted filters to reduce compression artifacts, which results in inferior results when processing images or videos with complicated patterns. 
Since compression artifacts reduction could be regarded as ill-posed inverse problems, prior knowledge plays an essential role in the process.
Several works propose to design effective priors to constrain the solution space \cite{zhang2016concolor,liu2016data,bredies2012total,chang2013reducing,song2020compressed,zhang2016low,zhang2013compression,liu2018graph}. Despite the higher performance they achieved, the representation abilities of handcrafted priors are still limited and usually fixed under different complex scenarios. 

Inspired by the advances in deep learning techniques, recent research efforts~\cite{zhang2022deep,jiang2021towards,vandal2018generating, RNANzhang2019residual,dncnnzhang2017beyond,li2017efficient,wang2019cfsnet,kim2020agarnet,chen2020adaptive,jin2020dual,qiu2020deep,zhang2020residual,fu2019jpeg,wang2016d3,guo2016building,chen2018dpw,zhang2018dmcnn,sun2020reduction,galteri2017deep,guo2017one,galteri2019deep,chen2018deep,zheng2019implicit,li2020learning,kim2019pseudo,ma2021reducing,ma2019residual,wang2017novel,liu2020comprehensive, ARCNNdong2015compression, mu2020graph,chen2021feature,zha2021triply,chen2021pre,liang2021swinir,cao2021video,wang2021uformer,zhang2019deep,dong2018denoising,jo2021rethinking,zhang2021plug,fu2021learning,fu2021model,liu2021underexposed,zhang2017learning,liu2019knowledge} propose to learn the inverse mapping of compressed images to original ones by deep neural networks. Despite their improved performances, these methods consider this process as an \textit{implicit} end-to-end learning process while neglecting the intrinsic attributes during compression, leading to unbalanced views of features when restoring images. For example, an artifacts reduction network that is well-trained on human face images shows an unsatisfactory performance when processing natural scene images. Hence, in this paper, we try to answer the following question: How to \textit{explicitly} model the intrinsic attributes during the compression process and thus guide the learning of artifacts reduction?

To address this issue, we make an insightful attempt to decouple the compression intrinsic attributes into two complementary aspects: 1) compression-insensitive features: the compressed images and original images should share the same semantic representations,~\eg, both belonging to the same \textit{person} class in~\figref{fig:motivation}. We thus resort to feature discriminator on high-level semantics by training a compression-insensitive auto-encoder beyond the basic restoration. In this manner, the semantic representations of compressed images and original ones could play similar effects when adapting to downstream high-level vision tasks, spanning over semantic segmentation, object detection, and recognition. 
2) compression-sensitive features: The local details, including blurriness and boundary sharpness, are easy to lose during compression. As it is hard to definite them using specific measurements, we thus propose to make the encoded features aware of quality factors, which reflect the degree of compression and are available in the compressed files. Based on these two decoupled aspects of features, our key idea is to utilize these features as learning guidance to adapt the decoded feature to various scenarios. Thereby, we introduce a new compression artifacts reduction framework, namely Dual Awareness Guidance Network (DAGN), which benefited from the successful decoupling of the semantic compression-insensitivity and low-level compression-sensitivity. Beyond the dual-awareness guidance, we also propose a cross-feature fusion module to incorporate the multi-scale semantic information from the compression-insensitive encoders. With the joint learning of the decoupling learning and dual awareness guidance, our proposed Dual Awareness Guidance Network is able to understand the low-level cues while also maintaining high-level semantic information during the image restoration process. To verify these contributions, we conduct experiments by two lines of results: 1) for visual quality, the proposed DAGN achieves superior performance against state-of-the-art methods on quantitative metrics, including PSNR and SSIM; 2) for downstream high-level vision tasks, we provide qualitative and quantitative results on object detection and semantic segmentation tasks after our artifacts reduction.

Our major contributions are as follows:
\begin{itemize}
    \item We revisit the compression process from a new perspective of decoupling learning and propose two compression awareness modules: the compression-insensitive modules for high-level semantics and the compression-sensitive module for compression details.
    \item We propose a Dual Awareness Guidance Network with a dual awareness guidance module to adapt the decoding process for various conditions and a cross-feature fusion module for multi-scale semantic enhancement.
    
    \item Extensive experiments are performed on synthetic and real compressed image datasets to validate the effectiveness and superiority of DAGN. We demonstrate that DAGN achieves favorable performance against state-of-the-art methods in both visual quality measurements and vision parsing tasks.
\end{itemize}

The remainder of this paper is organized as follows. Section \ref{sec:rw} reviews related works, and Section \ref{sec:pm} describes the proposed Dual Awareness Guidance Network. Experiments and analyses of the proposed method are presented in Section \ref{sec:exp}. Section \ref{sec:con} finally concludes this paper.

\section{Related Work}
\label{sec:rw}
Compression artifacts belong to image-dependent noises, and researchers have proposed numerous methods to reduce these artifacts\cite{liu2020comprehensive, bredies2012total}, primarily \emph{filtering-based methods}, \emph{model-based methods}, and \emph{learning-based methods}.

\subsection{Filtering-based methods}
Early methods pay attention to filter design. Zhai \emph{et al.} \cite{zhai2008efficient} construct a neighborhood related to the correlation in regions and utilize the filter in blocks. Yoo \emph{et al.} \cite{yoo2014post} propose to apply group-based filters to edge blocks and their similar ones.
Besides filtering in the pixel domain, Foi \emph{et al.} \cite{foi2007pointwise} propose to conduct filtering in the frequency domain. Overall, these early methods usually have limited performance when meeting images or videos with complicated patterns due to the limited representation capabilities of handcrafted filters.

\subsection{Prior-based methods}
Since artifacts reduction is ill-posed, it could be formulated as a maximum posterior problem, where prior knowledge could play an essential role in the process.
Some priors come from the compression process, such as quantization steps, and transformation coefficients could be used to constrain the artifacts reduction results \cite{zhang2016concolor,liu2016data}. 
Besides, researchers have utilized priors related to natural images, such as low rank \cite{zhang2016low}, non-local self-similarity \cite{zhang2013compression}, representation \cite{bredies2012total,chang2013reducing,song2020compressed}, and graph \cite{liu2018graph}, to reduce compression artifacts.
However, these methods usually need iterative computations, which are significantly time-consuming.

\subsection{Learning-based methods}
Learning-based methods aim to learn a mapping from compressed images to the original ones. Due to the powerful representation ability, deep neural networks have greatly improved compression artifacts reduction in recent years \cite{zhang2022deep,jiang2021towards,vandal2018generating,RNANzhang2019residual,dncnnzhang2017beyond,li2017efficient,wang2019cfsnet,kim2020agarnet,chen2020adaptive,jin2020dual,qiu2020deep,zhang2020residual,fu2019jpeg,wang2016d3,guo2016building,chen2018dpw,zhang2018dmcnn,sun2020reduction,galteri2017deep,guo2017one,galteri2019deep,chen2018deep,zheng2019implicit,li2020learning,kim2019pseudo,ma2021reducing,ma2019residual,wang2017novel,liu2020comprehensive,ARCNNdong2015compression, mu2020graph,chen2021feature,zha2021triply,chen2021pre,liang2021swinir,cao2021video,wang2021uformer,zhang2019deep,dong2018denoising,jo2021rethinking,zhang2021plug,fu2021learning,fu2021model,liu2021underexposed,zhang2017learning,liu2019knowledge}. The first deep neural networks-based method \cite{ARCNNdong2015compression} proposes a 4-layer network named ARCNN.
Motivated by the success of ResNet \cite{He2016Deepresnet} in image classification, deeper networks with residual learning are proposed for compression artifacts reduction, such as DnCNN \cite{dncnnzhang2017beyond}, RNAN \cite{RNANzhang2019residual}, and RDN \cite{zhang2020residual}.
Later, inspired by the superior performance of Transformer \cite{vaswani2017attention}, researchers also introduce Transformer for compression artifacts reduction \cite{chen2021pre, liang2021swinir,cao2021video, wang2021uformer}. Based on the standard Transformer architecture, Chen \emph{et al.} propose an image processing transformer for image restoration problems \cite{chen2021pre}.
Liang \emph{et al.} design SwinIR based on the Swin Transformer, which achieves better performance with fewer parameters.
Meanwhile, researchers propose to employ the generative adversarial network (GAN) \cite{radford2015unsupervised} for compression artifacts reduction to recover visually satisfying results from compressed images \cite{galteri2017deep, galteri2019deep, guo2017one}.

Besides approaches in the pixel domain, researchers propose many dual-domain learning-based methods\cite{wang2016d3,guo2017one,ehrlich2020quantization,zhang2018dmcnn,chen2018dpw,liu2018multi,sun2020reduction}, which exploit the knowledge of the frequency domain to reduce the artifacts.
Li \emph{et al.} \cite{li2020learning} propose a single network to process images compressed at all levels, which takes quantization steps and compressed images as input. 
To solve blind artifacts reduction where the quality factor (QF) is unavailable, Jiang \emph{et al.} \cite{jiang2021towards} and Kim \emph{et al.} \cite{kim2019pseudo} estimate QF by a network and then utilize the estimated QF to guide artifacts reduction.
Here, QF is an important index reflecting the compression degree of JPEG images, and a lower QF means less size is required, but more information is lost.
Recently, some researchers endeavor to combine prior-based and learning-based methods for compression artifacts reduction, where deep neural networks are employed to learn priors and perform as regularizers under the alternative minimization framework \cite{zhang2019deep,dong2018denoising,jo2021rethinking,zhang2021plug,fu2021learning,fu2021model,liu2021underexposed,zhang2017learning,liu2019knowledge,zha2021triply}. 
{
Dong \emph{et al.} \cite{dong2018denoising} adopt the half-quadratic splitting method to solve the minimization problem and designed a Denoising Prior Driven Deep Neural Network. Zhang \emph{et al.} \cite{zhang2019deep,zhang2021plug} set up a benchmark deep prior by training a highly flexible and effective CNN denoiser. Jo \emph{et al.} \cite{jo2021rethinking} analyze the prior by the notion of practical degrees of freedom to monitor the optimization progress. Fu \emph{et al.} \cite{fu2021model} utilize dictionary learning to model compression priors and design an iterative optimization algorithm using proximal operators. Subsequently, the priors used for this task are further differentiated. Fu \emph{et al.} \cite{fu2021learning} propose two prior terms for the image content and gradient, respectively. The content-relevant prior is formulated as an image-to-image regressor to perform as a deblocker from the pixel level. The gradient-relevant prior serves as a classifier to distinguish whether the image is compressed from the semantic level. Zha \emph{et al.} \cite{zha2021triply} propose a pair of triply complementary priors, namely, internal and external, shallow and deep, and non-local and local priors. They design a hybrid plug-and-play framework and a joint low-rank and deep image model based on the priors. Overall, These methods increase the interpretability of deep neural networks for compression artifacts reduction.
}

To summarize, learning-based approaches have significantly improved the performance of image compression artifacts reduction. However, the intrinsic attributes during the compression process are substantially neglected by prevailing methods. Hence, in this paper, we explore decoupling the compression image attributes from semantic-aware insensitive features and quality-aware sensitive features. 

\section{Approach}
\label{sec:pm}
\subsection{Sensitivity Decouple Learning}
\begin{figure}[t!]
    \centering
    \includegraphics[width=1.0\linewidth]{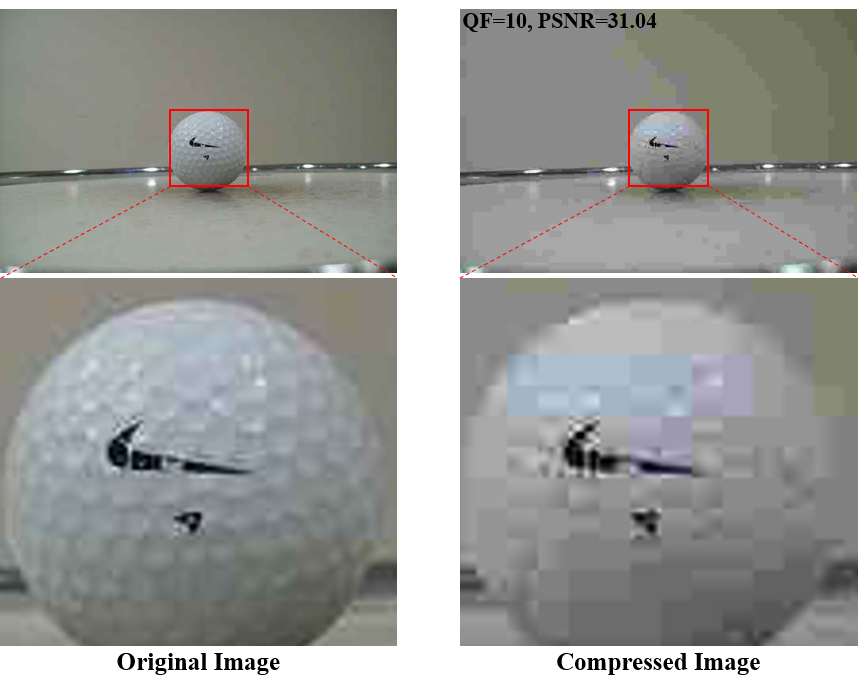}
    \caption{An original image and its JPEG compressed version with $\mathrm{QF}=10$ that shows the relationship between the intrinsic attributes and compression. Some intrinsic attributes are changed by JPEG compression, such as \emph{dimples}, while others are maintained, such as \emph{sphere}.}
    \label{fig:fig2}
\end{figure}
We first set up the basic notations used in this paper. Let $\mathbf{o}$ and $\mathbf{c}$ denote the original image and its compressed version. $\mathbf{A}_{\mathbf{o}}$ and $\mathbf{A}_{\mathbf{c}}$ be the corresponding intrinsic attributes, respectively.
Consider a given original image $\mathbf{o}$, and prevailing learning-based methods dedicate to learning a mapping function $\mathcal{F}$ from $\mathbf{c}$ to $\mathbf{o}$:
\begin{equation}
    \min_{\boldsymbol{\theta}_{\mathcal{F}}} \mathcal{L}(\mathcal{F}(\mathbf{c}; \boldsymbol{\theta}_{\mathcal{F}}), \mathbf{o}),
\end{equation}
where $\mathcal{F}(\cdot; \boldsymbol{\theta}_{\mathcal{F}})$ denotes the artifacts reduction network whose parameters are $\boldsymbol{\theta}_{\mathcal{F}}$; $\mathcal{L}$ is the evaluation metric, usually mean absolute error or mean square error.
This paper aims to obtain $\mathbf{A}_{\mathbf{c}}$ from $\mathbf{c}$ to promote the artifacts reduction process:
\begin{equation}
    \min_{\boldsymbol{\theta}_{\mathcal{F}}} \mathcal{L}(\mathcal{F}(\mathbf{c}; \boldsymbol{\theta}_{\mathcal{F}}|\mathbf{A}_\mathbf{c}), \mathbf{o}).
\end{equation}

During the compression process, it is known that some certain attributes in $\mathbf{A}_{\mathbf{c}}$ are usually broken or degraded, while others are substantially retained. Therefore, we argue that these two components contain different cues: the unbroken part holds some high-level semantic information, while the broken part reflects the degree of compression/quality. In~\figref{fig:fig2}, we could obtain \emph{sphere} and \emph{block patterns on the surface} from the compressed image. Comparing the compressed image with the original one, we could observe that \emph{sphere} is not broken by compression, while \emph{block patterns on the surface} are broken during compression. In other words, we could obtain \emph{sphere} and \emph{dimples} from the original image. \emph{sphere} exists in the compressed image, while \emph{dimples} do not exist and are replaced by \emph{block patterns on the surface}. 
Therefore, considering the maintenance during compression, the intrinsic attributes of compressed images can be divided into two complementary components,~\ie, compression-insensitive features $\mathbf{F}_{cif}$ and compression-sensitive features $\mathbf{F}_{csf}$. $\mathbf{F}_{cif}$ shares similar high-level semantic information with the original image, while $\mathbf{F}_{csf}$ holds the information related to the low-level qualities, which are highly related to the Quality Factors (QF).
To decouple these two features, $\mathbf{F}_{cif}$ and $\mathbf{F}_{csf}$ from $\mathbf{c}$, we employ a common approach that automatically learns features from unlabeled data, auto-encoders  \cite{bengio2009learning,ng2011sparse,liou2014autoencoder,meng2017relational,tschannen2018recent}.
{
Bengio \emph{et al.} \cite{bengio2009learning} summarize the formulation of auto-encoders and discuss their connection to Restricted Boltzmann Machines. 
Ng \emph{et al.} \cite{ng2011sparse} impose a sparsity constraint on the hidden units of the auto-encoders and explore the structure in the data.
Meng \emph{et al.} \cite{meng2017relational} propose a relation auto-encoder that could extract high-level features based on both data itself and their relationships.
Tschannen \emph{et al.} \cite{tschannen2018recent} review representation learning with a focus on auto-encoder-based models, which could be referred to get a quick understanding of auto-encoder. 
}
We define them in a former way: one typical auto-encoder could be considered to include two modules: one encoder $\mathcal{E}(\cdot; \boldsymbol{\theta}_{\mathcal{E}})$ to embed the image into feature representations; and one decoder $\mathcal{D}(\cdot; \boldsymbol{\theta}_{\mathcal{D}})$ to reconstruct and estimate the uncompressed images from the representations, which could be denoted as
$\mathcal{D}(\mathcal{E}(\cdot; \boldsymbol{\theta}_{\mathcal{E}});\boldsymbol{\theta}_{\mathcal{D}})$.

\subsubsection{Compression-Insensitive Auto-Encoder}
\begin{figure}[t!]
    \centering
    \includegraphics[width=1.00\linewidth]{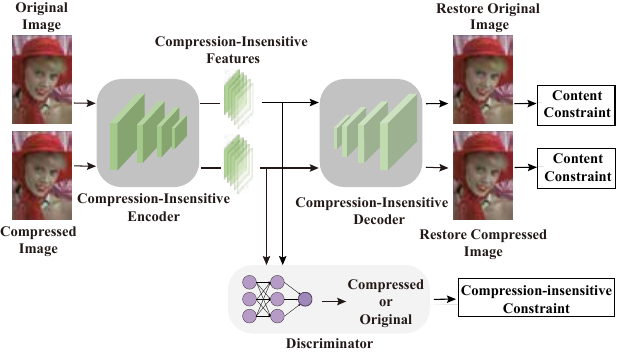}
    \caption{The training of compression-insensitive auto-encoder. The discriminator takes the features learned by the compression-insensitive encoder as the input and generates the probability that the features are from original images. By training the auto-encoder to fool the differentiable discriminator network, we obtain the compression-insensitive encoder that could extract compression-insensitive features from images.}
    \label{fig:IE}
\end{figure}
\begin{figure}[t!]
    \centering
    \includegraphics[width=1.00\linewidth]{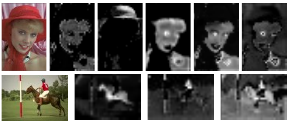}
    \caption{Visualization of the compression-insensitive features learned from \emph{womanhat} in LIVE1 and \emph{206062} in BSD500.}
    \label{fig:ief}
\end{figure}
Our key idea for constructing compression-insensitive features is to maintain the semantic consistency of original and compressed images through adversarial training. 
We denote the auto-encoder to extract compression-insensitive features as $\mathcal{E}_{ci}(\cdot; \boldsymbol{\theta}^{ci}_\mathcal{E})$ and $\mathcal{D}_{ci}(\cdot; \boldsymbol{\theta}^{ci}_{\mathcal{D}})$. 
{
Specifically, the former is a modified ResNet-50 encoder whose average-pooling layers are removed, and the last stride 2 convolution is replaced with stride 1. The latter is a 15-layer fully convolutional decoder, consisting of 5 convolutional layers followed by alternating convolution and convolution transpose layers.
For compression-insensitive features, we employ a discriminator network $\mathcal{J}(\cdot; \boldsymbol{\theta}_{\mathcal{J}})$ which is a $2048\times 2048 \times 1$ MLP. 
}
In other words, if the high-level features of original images and compressed ones could not be distinguished, their features would be highly consistent. $\mathcal{J}(\cdot; \boldsymbol{\theta}_{\mathcal{J}})$ takes the features learned by $\mathcal{E}_{ci}(\cdot; \boldsymbol{\theta}^{ci}_\mathcal{E})$ as input and generates the probability that the features are from original images. During training, we optimize $\boldsymbol{\theta}_{\mathcal{J}}$ in an alternating manner along with the compression-insensitive encoder and decoder. The updating of  ${\boldsymbol{\theta}^{ci}_\mathcal{E},\boldsymbol{\theta}^{ci}_\mathcal{D}}$ can be formally represented as:
\begin{align}
    \boldsymbol{\theta}^{ci}_\mathcal{E} &\leftarrow \boldsymbol{\theta}^{ci}_\mathcal{E} - lr \nabla_{\boldsymbol{\theta}^{ci}_\mathcal{E}} (\mathcal{L}^{ci}_{cc} + \lambda_{ci}\mathcal{L}_{cic}), \notag\\
    \boldsymbol{\theta}^{ci}_\mathcal{D} &\leftarrow \boldsymbol{\theta}^{ci}_\mathcal{D} - lr \nabla_{\boldsymbol{\boldsymbol{\theta}^{ci}_\mathcal{D}}} (\mathcal{L}^{ci}_{cc} + \lambda_{ci}\mathcal{L}_{cic}),
    \label{eq:ie_train}
\end{align}
and the one of $\boldsymbol{\theta}_{\mathcal{J}}$:
\begin{align}
    \boldsymbol{\theta}_\mathcal{J} \leftarrow \boldsymbol{\theta}_\mathcal{J} - lr \nabla_{\boldsymbol{\theta}_\mathcal{J}} (- \mathcal{L}_{cic}),
\end{align}
where $\lambda_{ci}$ is a penalty factor. $lr$ denotes the learning rate. $\mathcal{L}^{ci}_{cc}$ and $\mathcal{L}_{cic}$ denote the content constraint and the compression-insensitive constraint (shown in~\figref{fig:IE}):
\begin{align}
    \mathcal{L}^{ci}_{cc} &= \frac{1}{2N} \sum^{N}_{i=1} \underbrace{\left\|  \mathcal{D}_{ci}(\mathcal{E}_{ci}(\mathbf{c}_{i}; \boldsymbol{\theta}^{ci}_\mathcal{E}); \boldsymbol{\theta}^{ci}_{\mathcal{D}}) - \mathbf{c}_{i} \right\|_{1}}_{\text{content constraint on }\mathbf{c}_{i}} \notag\\
    & + \underbrace{\left\| \mathcal{D}_{ci}(\mathcal{E}_{ci}(\mathbf{o}_{i}; \boldsymbol{\theta}^{ci}_\mathcal{E}); \boldsymbol{\theta}^{ci}_{\mathcal{D}}) - \mathbf{o}_{i} \right\|_{1}}_{\text{content constraint on }\mathbf{o}_{i}},
    \label{eq:lcc1}
\end{align}
\begin{align}
    \mathcal{L}_{cic} = \frac{1}{2N} \sum^{N}_{i=1}     \underbrace{- \log \mathcal{J}(\mathcal{E}_{ci}(\mathbf{c}_{i}; \boldsymbol{\theta}^{ci}_\mathcal{E}); \boldsymbol{\theta}_{\mathcal{J}})}_{\text{compression-insensitive constraint on } \mathbf{c}_{i}} \notag\\
    \underbrace{+ \log \mathcal{J}(\mathcal{E}_{ci}(\mathbf{o}_{i}; \boldsymbol{\theta}^{ci}_\mathcal{E}); \boldsymbol{\theta}_{\mathcal{J}})}_{\text{compression-insensitive constraint on }\mathbf{o}_{i}}.
    \label{eq:lcic}
\end{align}
where $N$ is the batch size during training. The idea behind this is that it allows one to train an auto-encoder with the goal of fooling a differentiable discriminator network that is trained to distinguish the features of compressed images from that of original images.~\figref{fig:ief} exhibits the visualization of the compression-insensitive features learned from two samples.

\subsubsection{Compression-Sensitive Auto-Encoder}
\begin{figure}[t!]
    \centering
    \includegraphics[width=1.00\linewidth]{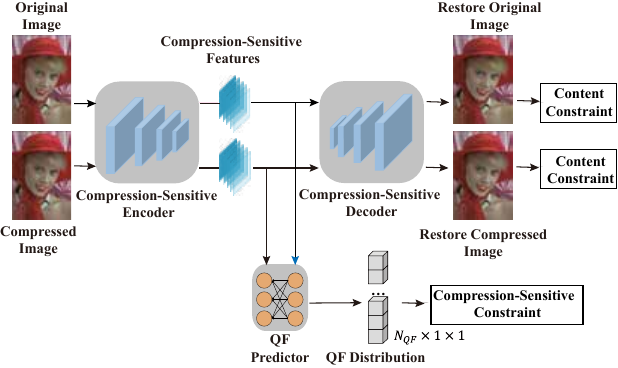}
    \caption{The training of compression-sensitive auto-encoder. The QF predictor takes the features learned by the compression-sensitive encoder as the input and generates the QF distribution. By training the auto-encoder with the goal of recognizing the QF, we obtain a compression-sensitive encoder that could extract compression-insensitive features from images.}
    \label{fig:SE}
\end{figure}
\begin{figure}[t!]
    \centering
    \includegraphics[width=1.00\linewidth]{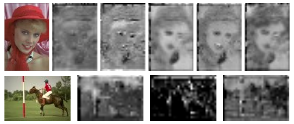}
    \caption{Visualization of the compression-sensitive features learned from \emph{womanhat} in LIVE1 and \emph{206062} in BSD500.}
    \label{fig:sef}
\end{figure}
As aforementioned, here we resort to the quality factor (QF) to represent the compression degree of compressed images and regularize features to be aware of compression information. Therefore, the features tend to understand low-level details that are sensitive during compression.
{
To this end, we construct an encoder $\mathcal{E}_{cs}(\cdot; \boldsymbol{\theta}^{cs}_{\mathcal{E}})$ to extract compression-sensitive features and $\mathcal{D}_{cs}(\cdot; \boldsymbol{\theta}^{cs}_{\mathcal{D}})$ for decoding, which share the same architecture with the compression-insensitive auto-encoder.
}
To extract compression-sensitive features, we rely on an auxiliary compression degree predictor $\mathcal{P}_{qf}(\cdot; \boldsymbol{\theta}^{qf}_{\mathcal{P}})$ to $\mathcal{E}_{cs}(\cdot; \boldsymbol{\theta}^{cs}_{\mathcal{E}})$ and $\mathcal{D}_{cs}(\cdot; \boldsymbol{\theta}^{cs}_{\mathcal{D}})$.
{
The predictor is a $2048\times2048\times100$ MLP, which takes the features extracted by $\mathcal{E}_{cs}(\cdot; \boldsymbol{\theta}^{cs}_{\mathcal{E}})$ as input and outputs the compression degree.
}
Training an auto-encoder with the goal of predicting the QF from the features could supervise the encoder to extract compression-sensitive features. In training, we update $\boldsymbol{\theta}^{cs}_{\mathcal{E}},\boldsymbol{\theta}^{cs}_{\mathcal{D}},\boldsymbol{\theta}^{qf}_{\mathcal{P}}$ by:
\begin{align}
    \boldsymbol{\theta}^{cs}_{\mathcal{E}} &\leftarrow \boldsymbol{\theta}^{cs}_{\mathcal{E}} - lr \nabla_{\boldsymbol{\theta}^{cs}_{\mathcal{E}}} (\mathcal{L}^{cs}_{cc} + \lambda_{cs} \mathcal{L}_{csc}), \notag\\
    \boldsymbol{\theta}^{cs}_{\mathcal{D}} &\leftarrow \boldsymbol{\theta}^{cs}_{\mathcal{D}} - lr \nabla_{\boldsymbol{\theta}^{cs}_{\mathcal{D}}} (\mathcal{L}^{cs}_{cc} + \lambda_{cs} \mathcal{L}_{csc}), \notag\\
    \boldsymbol{\theta}^{qf}_{\mathcal{P}} &\leftarrow \boldsymbol{\theta}^{qf}_{\mathcal{P}} - lr \nabla_{\boldsymbol{\theta}^{qf}_{\mathcal{P}}} (\mathcal{L}^{cs}_{cc} + \lambda_{cs} \mathcal{L}_{csc}),
    \label{eq:se_train}
\end{align}
where $\lambda_{cs}$ is a penalty factor, $\mathcal{L}^{cs}_{cc}$ and $\mathcal{L}_{csc}$ denote the content constraint and the compression-sensitive constraint (in~\figref{fig:SE}):
\begin{align}
    \mathcal{L}^{cs}_{cc} &= \frac{1}{2N} \sum^{N}_{i=1} \underbrace{\left\|  \mathcal{D}_{cs}(\mathcal{E}_{cs}(\mathbf{c}_{i}; \boldsymbol{\theta}^{cs}_{\mathcal{E}}); \boldsymbol{\theta}^{cs}_{\mathcal{D}}) - \mathbf{c}_{i} \right\|_{1}}_{\text{content constraint on }\mathbf{c}_{i}} \notag\\
    & + \underbrace{\left\| \mathcal{D}_{cs}(\mathcal{E}_{cs}(\mathbf{o}_{i}; \boldsymbol{\theta}^{cs}_{\mathcal{E}}); \boldsymbol{\theta}^{cs}_{\mathcal{D}}) - \mathbf{o}_{i} \right\|_{1}}_{\text{content constraint on }\mathbf{o}_{i}},
    \label{eq:lcc2}
\end{align}
\begin{align}
    \mathcal{L}_{csc} &= \frac{1}{2N} \sum^{N}_{i=1} \underbrace{\left\| \mathcal{P}_{qf}(\mathcal{E}_{cs}(\mathbf{c}_{i}; \boldsymbol{\theta}^{cs}_{\mathcal{E}}); \boldsymbol{\theta}^{qf}_{\mathcal{P}}) - \bs{QF}_{c}^{i} \right\|_{1}}_{\text{compression-sensitive constraint on }\mathbf{c}_{i}}  \notag\\
    & + \underbrace{\left \|\mathcal{P}_{qf}(\mathcal{E}_{cs}(\mathbf{o}_{i}; \boldsymbol{\theta}^{cs}_{\mathcal{E}}); \boldsymbol{\theta}^{qf}_{\mathcal{P}}) - \bs{QF}_{\mathbf{o}} \right\|_{1}}_{\text{compression-sensitive constraint on }\mathbf{o}_{i}},
    \label{eq:lcsc}
\end{align}
where $\bs{QF}$ denotes the one-hot encodings of $100$ QF classes. During this standard classification process, the features are aware of the quality factors and thus sensitive to compression degrees. 
{
~\figref{fig:sef} visualizes the compression-sensitive features learned from two representative samples.
}

\subsection{Dual Awareness Guidance Network}
\begin{figure*}[t!]
    \centering
    \includegraphics[width=1.00\linewidth]{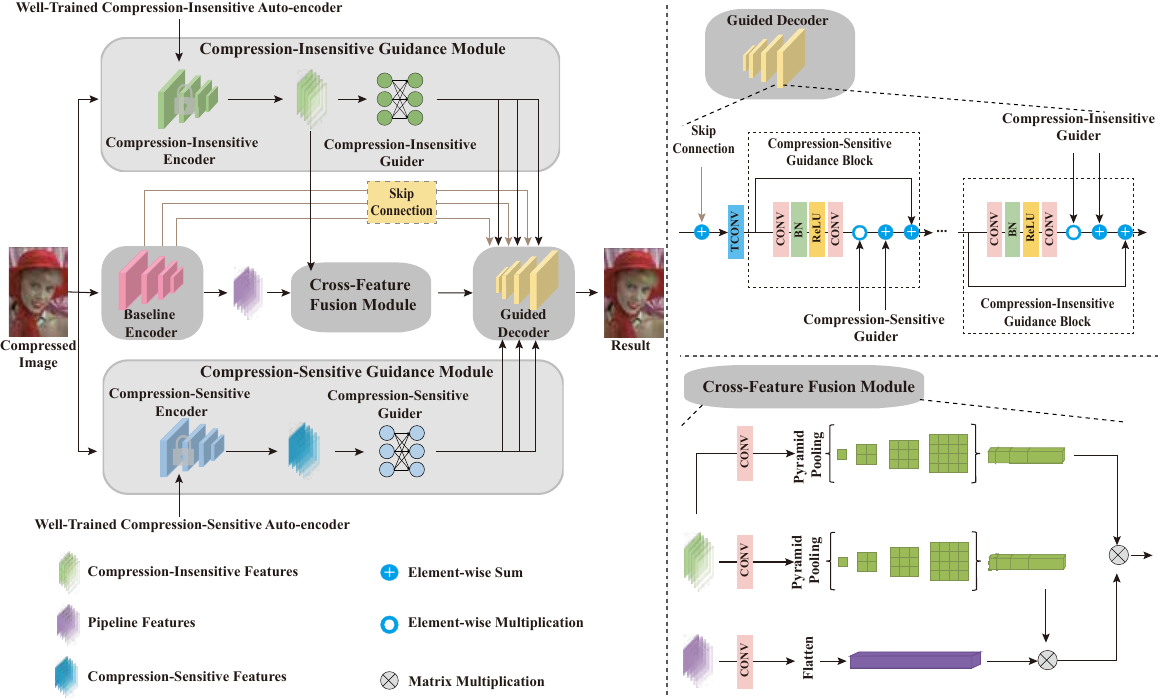}
    \caption{The architecture of DAGN for compression artifacts reduction. DAGN utilizes the decoupled sensitive and insensitive features as guidance for the feature decoding process. The compression-insensitive guidance module and compression-sensitive guidance module aim to extract compression-insensitive features and compression-sensitive features to guide the compression artifacts reduction process. Cross-feature fusion module aims to fuse compression-insensitive features into the artifacts reduction process. }
    \label{fig:DAGN}
\end{figure*}
Following the encoding-decoding fashion in compression artifacts reduction \cite{jiang2021towards,zhang2021plug,dong2018denoising,liang2021swinir,wang2021uformer}, our DAGN is also composed of one base encoder and one guided decoder. To build connections and guidance during learning, we introduce three insightful modules, \emph{i.e.}, the compression-insensitive guidance module, compression-sensitive guidance module, and cross-feature fusion module.~\figref{fig:DAGN} shows the whole architecture of DAGN, which takes compression images as input and outputs the artifacts reduction results.
\subsubsection{Base Encoder}
The base encoder involves four spatial scales, each of which owns a skip connection to the guided decoder. In each scale, we employ four residual blocks, which are composed of two $3 \times 3$ convolution layers with ReLU activation and batch normalization in the middle. In the tail of the first three scales, convolutions with $2 \times 2$ strides are adopted to downscale the features. We set the channel numbers of features in the first to the last scale to 64, 128, 256, and 512, respectively.

\subsubsection{Compression-Insensitive Guidance}
The compression-insensitive guidance module aims to extract compression-insensitive features and utilize them to guide the artifacts reduction process. The compression-insensitive guidance takes well-trained insensitive features from the auto-encoder and then learns parameters for transforming feature representations in the decoding process. We employ a three-layer Multi-Layer Perceptron (MLP) as a compression-insensitive guider to learn these transformation features. The first two layers generate shared intermediate conditions, while the last layer generates different parameters $(\boldsymbol{\beta}_{i}, \boldsymbol{\gamma}_{i}), i=1,2,3$ for the three dual awareness guidance blocks in the guided decoder. 

\subsubsection{Compression-Sensitive Guidance} 
Analogous to the insensitive guidance, our goal is to extract compression-sensitive features and utilize them to guide the artifacts reduction process. We first use the compression-sensitive guider to encode compression-insensitive features, then learn parameters $(\boldsymbol{\epsilon}_{i}, \boldsymbol{\eta}_{i}), i=1,2,3$ that could be fused into the dual awareness guidance blocks in the guided decoder.
\subsubsection{Cross-Feature Fusion Module}
{
Compression-insensitive features represent shared semantic information between compressed and original images. This information should be evident in the results of reducing artifacts.
}
To maintain the consistency of compression-insensitive features between original images and artifacts reduction results, we propose the cross-feature fusion module (CFM), which
fuses compression-insensitive features into the artifacts reduction baseline.

The compression-insensitive features $\mathbf{F}_{cif} \in \mathbb{R}^{C_{c}\times H_{c} \times W_{c}}$ and base features $\mathbf{F}_{bf} \in \mathbb{R}^{C_{b}\times H_{b} \times W_{b}}$ are in different feature spaces. Thus, to fuse them with enhanced receptive fields, CFM firstly conducts three $1 \times 1$ convolutions to transform $\mathbf{F}_{cif}$ and $\mathbf{F}_{bf}$ to features with the same channel number $C_{b}$, where $\mathcal{C}^{1}_{1}(\mathbf{F}_{cif}),\mathcal{C}^{1}_{2}(\mathbf{F}_{cif}) \in \mathbb{R}^{C_{b}\times H_{c} \times W_{c}}$, $\mathcal{C}^{1}_{3}(\mathbf{F}_{bf}) \in \mathbb{R}^{C_{b}\times H_{b} \times W_{b}}$, 
and $\{\mathcal{C}^{1}_{i}(\cdot)\}_{i=1}^{3} $ denote different $1 \times 1$ convolutions.
Additionally, a pyramid pooling \cite{lazebnik2006beyond} operation $\bs{Pool}$ that contains four pooling layers with different output sizes ($\{k\times k\}_{k=1,2,4,8}$) in parallel is conducted to sample $\mathcal{C}^{1}_{1}(\mathbf{F}_{cif})$ and $\mathcal{C}^{1}_{2}(\mathbf{F}_{cif})$. Here $\bs{Pool}(\mathcal{C}^{1}_{1}(\mathbf{F}_{cif})), \bs{Pool}(\mathcal{C}^{1}_{2}(\mathbf{F}_{cif})) \in \mathbb{R}^{C_{b}\times (\sum_{n=1,2,4,8} n^2)}$. Our proposed CFM has the following form:
\begin{align}
    \mathcal{M}_{cf}(\mathbf{F}_{cif}, \mathbf{F}_{bf}) = & \mathcal{C}^{1}_{3}(\mathbf{F}_{bf}) \times \bs{Pool}(\mathcal{C}^{1}_{1}(\mathbf{F}_{cif}))  \\ \notag
    & \times \bs{Pool}(\mathcal{C}^{1}_{2}(\mathbf{F}_{cif}))^{T} + \mathbf{F}_{bf}^{T}, 
\end{align}
where $\times$ denotes the matrix multiplication operation, and $\mathcal{C}^{1}_{3}(\mathbf{F}_{bf}) \times \bs{Pool}(\mathcal{C}^{1}_{1}(\mathbf{F}_{cif}))$ is an approximate similarity matrix between $\mathbf{F}_{cif}$ and $\mathbf{F}_{bf}$.

\subsubsection{Guided Decoder Transformation}
The guided decoder transformation aims to utilize the decoupled sensitive and insensitive features as learning guidance to transform the encoded representations, which helps to build an adaptive decoding process under different scenarios.
Our guided decoder involves four scales, each of which receives a skip connection to the base encoder. The first scale consists of four residual blocks comprising two $3 \times 3$ convolution layers with ReLU activation and batch normalization in the middle. The last three scales employ Dual Awareness Guidance Blocks (DAGB). As in~\figref{fig:DAGN}, DAGB receives the skip connection from the corresponding scale in the base encoder, and the parameters learned from the compression-sensitive guider and compression-insensitive guider (\emph{i.e.} $(\boldsymbol{\beta}_{i}, \boldsymbol{\gamma}_{i}), (\boldsymbol{\epsilon}_{i}, \boldsymbol{\eta}_{i})$). 
With the embedding dimension of $D$, thus $\boldsymbol{\beta}_{i}, \boldsymbol{\gamma}_{i}, \boldsymbol{\epsilon}_{i}, \boldsymbol{\eta}_{i} \in \mathbb{R}^{{\frac{D}{2^{i}}}\times 1}$ represent the feature scale factors and transformation biases.
Specifically, DAGN adopts $2 \times 2$ transpose convolutions for upscaling operations. Then, four compression-sensitive guidance blocks and four compression-insensitive guidance blocks are utilized to guide the artifacts reduction process by applying transformations to intermediate features.

Let $\mathbf{F}^{in}_{dagb}$ be the input of one DAGB in the guided decoder, and $\mathbf{F}^{sc}_{dagb}$ denote the corresponding skip connection. The output of the \emph{i-th} DAGB could be written as:
\begin{equation}
    \mathcal{B}_{dagb}(\mathbf{F}^{in}_{dagb}, \mathbf{F}^{sc}_{dagb}) = \mathcal{B}_{cig}^{4}(\mathcal{B}_{csg}^{4}(\mathcal{T}_{2}(\mathbf{F}^{in}_{dagb} + \mathbf{F}^{sc}_{dagb}))),
\end{equation}
where $\mathcal{T}_{2}$ denotes the transpose convolution layers; $\mathcal{B}_{cig}(\cdot)$ and $\mathcal{B}_{csg}(\cdot)$ denote the compression-insensitive guidance block and compression-sensitive guidance block, respectively.
\begin{equation}
    \mathcal{B}_{cig}(\mathbf{F}^{in}_{cig}) = \mathbf{F}^{in}_{cig} +  \boldsymbol{\beta}_{i} \odot \mathcal{C}^{3}_{1}(\bs{RNorm}(\mathcal{C}^{3}_{2}(\mathbf{F}^{in}_{cig}))) + \boldsymbol{\gamma}_{i},
\end{equation}
\begin{equation}
    \mathcal{B}_{csg}(\mathbf{F}^{in}_{csg}) = \mathbf{F}^{in}_{csg} +  \boldsymbol{\epsilon}_{i} \odot \mathcal{C}^{3}_{3}(\bs{RNorm}(\mathcal{C}^{3}_{4}(\mathbf{F}^{in}_{csg}))) + \boldsymbol{\eta}_{i}.
\end{equation}
Here $\mathbf{F}^{in}_{cig}$ and $\mathbf{F}^{in}_{csg}$ denote the input of the compression-insensitive guidance block and the compression-sensitive guidance block, respectively. $\odot$ is channel-wise multiplication; $\{\mathcal{C}^{3}_{i}(\cdot)\}_{i=1}^{4}$ are different $3 \times 3$ convolutions; $\bs{RNorm}$ denote ReLU activation and batch normalization.

During training, given a batch of training samples, we freeze the parameters of the compression-insensitive encoder and compression-sensitive encoder, and optimize the rest parameters by minimizing the $\mathcal{L}_{1}$ pixel loss:
\begin{equation}
    \mathcal{L}_{1} = \frac{1}{N} \sum^{N}_{i=1} \left\|\mathcal{F}_{dagn}(\mathbf{c}_{i}; \hat{\boldsymbol{\theta}}^{ci}_{\mathcal{E}}, \hat{\boldsymbol{\theta}}^{cs}_{\mathcal{E}}, \boldsymbol{\theta}_{res}) - \mathbf{o}_{i} \right\|_{1},
    \label{eq:l1}
\end{equation}
\begin{equation}
    \boldsymbol{\theta}_{res} \leftarrow \boldsymbol{\theta}_{res} - \nabla_{\boldsymbol{\theta}_{res}} \mathcal{L}_{1}.
    \label{eq:dagn_train}
\end{equation}
Here $\mathcal{F}_{dagn}(\cdot; \hat{\boldsymbol{\theta}}^{ci}_{\mathcal{E}}, \hat{\boldsymbol{\theta}}^{cs}_{\mathcal{E}}, \boldsymbol{\theta}_{res})$ denotes the DAGN with frozen compression-insensitive encoder and compression-sensitive encoder, where $\hat{\boldsymbol{\theta}}^{ci}_{\mathcal{E}}$ and $\hat{\boldsymbol{\theta}}^{cs}_{\mathcal{E}}$ are the parameters of well-trained compression-insensitive encoder and compression-sensitive encoder; $\boldsymbol{\theta}_{res}$ denotes the rest parameters of DAGN (\emph{i.e.} the parameters of compression-insensitive guider, compression-sensitive guider, base encoder, cross-feature fusion module, and guided decoder). The overall training scheme is summarized in Algorithm \ref{algorithm1}.
\begin{algorithm}[t!]
	\small
	\caption{Algorithm of DAGN Training.}
	\label{algorithm1}
	\begin{algorithmic}[1]
		\State {\bf Input:} Original and compressed image pairs: $\mathbb{D}=<\mathbf{c}_{i},\mathbf{o}_{i}>$, hyperparameters $lr,ite_1, ite_2, \lambda_{ci}, \lambda_{cs}$.
		\State {\bf Output:} The parameters of DAGN: $\hat{\boldsymbol{\theta}}^{ci}_{\mathcal{E}}$, $\hat{\boldsymbol{\theta}}^{cs}_{\mathcal{E}}$, and  $\hat{\boldsymbol{\theta}}_{res}$.
		{\color{blue} \Statex {/* Optimization of  $\boldsymbol{\theta}^{ci}_{\mathcal{E}},\boldsymbol{\theta}^{cs}_{\mathcal{E}}$\ */} }
		\State $ite \leftarrow 0$
		\For {$ite < ite_1$}
		\State Sample a mini-batch $<\mathbf{c}_{i},\mathbf{o}_{i}> \subset \mathbb{D}, i=1,2,...,N$
		\State Calculate $\mathcal{L}^{ci}_{cc}$ and $\mathcal{L}_{cic}$ by Equation (\ref{eq:lcc1}) and (\ref{eq:lcic})
        \State Update $\boldsymbol{\theta}^{ci}_{\mathcal{E}}, \boldsymbol{\theta}^{ci}_{\mathcal{D}}, \boldsymbol{\theta}_{\mathcal{J}}$ by Equation (\ref{eq:ie_train})
        \State Calculate $\mathcal{L}^{cs}_{cc}$ and $\mathcal{L}_{csc}$ by Equation (\ref{eq:lcc2}) and (\ref{eq:lcsc})
        \State Update $\boldsymbol{\theta}^{cs}_{\mathcal{E}}, \boldsymbol{\theta}^{cs}_{\mathcal{D}}, \boldsymbol{\theta}^{qf}_{\mathcal{P}}$ by Equation (\ref{eq:se_train})
        \State $ite \leftarrow ite + 1$
		\EndFor
		\State $\hat{\boldsymbol{\theta}}^{ci}_{\mathcal{E}},\hat{\boldsymbol{\theta}}^{cs}_{\mathcal{E}} \leftarrow \boldsymbol{\theta}^{ci}_{\mathcal{E}},\boldsymbol{\theta}^{cs}_{\mathcal{E}}$ 
		{\color{blue} \Statex {/* Optimization of $\boldsymbol{\theta}_{res}$\ */}}
		\State $ite \leftarrow 0$
		\For {$ite < ite_2$}
		\State Sample a mini-batch $<\mathbf{c}_{i},\mathbf{o}_{i}> \subset \mathbb{D}, i=1,2,...,N$
		\State Calculate $\mathcal{L}_{1}$ by Equation (\ref{eq:l1})
		\State Update $\boldsymbol{\theta}_{res}$ by Equation (\ref{eq:dagn_train})
		\State $ite \leftarrow ite + 1$
		\EndFor
		\State $\hat{\boldsymbol{\theta}}_{res} \leftarrow \boldsymbol{\theta}_{res}$
	\end{algorithmic}  
\end{algorithm}

\begin{table*}[t]
\renewcommand{\arraystretch}{1.25}
\caption{Quantitative comparisons on LIVE1 and BSD500. Please note that the multi-model methods train a specific model for each quality factor, while the single-model methods train one model for all quality factors.}
\centering
\label{tab:quantitative_comparison}
\begin{tabular}{|c|c|c|c|cccc|ccc|}
\hline
\multirow{2}{*}{Dataset} & \multirow{2}{*}{QF} & \multirow{2}{*}{Metric} &       & \multicolumn{4}{c|}{Multi   Models}                                                                      & \multicolumn{3}{c|}{Single   Model}                                                     \\ \cline{4-11} 
                         &                     &                         & JPEG  & \multicolumn{1}{c|}{ARCNN \cite{ARCNNdong2015compression}} & \multicolumn{1}{c|}{DMCNN \cite{zhang2018dmcnn}} & \multicolumn{1}{c|}{IPT \cite{chen2021pre}}         & SWINIR \cite{liang2021swinir}      & \multicolumn{1}{c|}{QGAC \cite{ehrlich2020quantization}}           & \multicolumn{1}{c|}{FBCNN \cite{jiang2021towards}}       & Ours           \\ \hline
\multirow{12}{*}{LIVE1}  & \multirow{3}{*}{10} & PSNR                    & 25.69 & \multicolumn{1}{c|}{26.43} & \multicolumn{1}{c|}{27.18} & \multicolumn{1}{c|}{27.37}       & 27.45       & \multicolumn{1}{c|}{27.62}          & \multicolumn{1}{c|}{{\ul 27.77}} & \textbf{27.95} \\ \cline{3-11} 
                         &                     & SSIM                    & 0.743 & \multicolumn{1}{c|}{0.770} & \multicolumn{1}{c|}{0.802} & \multicolumn{1}{c|}{0.799}       & 0.796       & \multicolumn{1}{c|}{{\ul 0.804}}    & \multicolumn{1}{c|}{0.803}       & \textbf{0.807} \\ \cline{3-11} 
                         &                     & PSNR-B                  & 24.20 & \multicolumn{1}{c|}{26.32} & \multicolumn{1}{c|}{27.03} & \multicolumn{1}{c|}{27.34}       & 27.23       & \multicolumn{1}{c|}{27.43}          & \multicolumn{1}{c|}{{\ul 27.51}} & \textbf{27.70} \\ \cline{2-11} 
                         & \multirow{3}{*}{20} & PSNR                    & 28.06 & \multicolumn{1}{c|}{28.86} & \multicolumn{1}{c|}{30.01} & \multicolumn{1}{c|}{29.99}       & 29.93       & \multicolumn{1}{c|}{29.88}          & \multicolumn{1}{c|}{{\ul 30.11}} & \textbf{30.27} \\ \cline{3-11} 
                         &                     & SSIM                    & 0.826 & \multicolumn{1}{c|}{0.843} & \multicolumn{1}{c|}{0.850} & \multicolumn{1}{c|}{0.859}       & 0.866       & \multicolumn{1}{c|}{{\ul 0.868}}    & \multicolumn{1}{c|}{{\ul 0.868}} & \textbf{0.873} \\ \cline{3-11} 
                         &                     & PSNR-B                  & 26.49 & \multicolumn{1}{c|}{28.61} & \multicolumn{1}{c|}{29.08} & \multicolumn{1}{c|}{29.51}       & 29.56       & \multicolumn{1}{c|}{29.56}          & \multicolumn{1}{c|}{{\ul 29.70}} & \textbf{29.87} \\ \cline{2-11} 
                         & \multirow{3}{*}{30} & PSNR                    & 29.37 & \multicolumn{1}{c|}{29.96} & \multicolumn{1}{c|}{-}     & \multicolumn{1}{c|}{31.04}       & 31.12       & \multicolumn{1}{c|}{31.17}          & \multicolumn{1}{c|}{{\ul 31.43}} & \textbf{31.60} \\ \cline{3-11} 
                         &                     & SSIM                    & 0.861 & \multicolumn{1}{c|}{0.879} & \multicolumn{1}{c|}{-}     & \multicolumn{1}{c|}{0.888}       & 0.896       & \multicolumn{1}{c|}{0.896}          & \multicolumn{1}{c|}{{\ul 0.897}} & \textbf{0.900} \\ \cline{3-11} 
                         &                     & PSNR-B                  & 27.84 & \multicolumn{1}{c|}{28.79} & \multicolumn{1}{c|}{-}     & \multicolumn{1}{c|}{30.79}       & 30.84       & \multicolumn{1}{c|}{30.77}          & \multicolumn{1}{c|}{{\ul 30.92}} & \textbf{31.10} \\ \cline{2-11} 
                         & \multirow{3}{*}{40} & PSNR                    & 30.28 & \multicolumn{1}{c|}{30.78} & \multicolumn{1}{c|}{-}     & \multicolumn{1}{c|}{32.09}       & 32.29       & \multicolumn{1}{c|}{32.05}          & \multicolumn{1}{c|}{{\ul 32.34}} & \textbf{32.50} \\ \cline{3-11} 
                         &                     & SSIM                    & 0.882 & \multicolumn{1}{c|}{0.896} & \multicolumn{1}{c|}{-}     & \multicolumn{1}{c|}{0.903}       & 0.912       & \multicolumn{1}{c|}{0.912}          & \multicolumn{1}{c|}{{\ul 0.913}} & \textbf{0.916} \\ \cline{3-11} 
                         &                     & PSNR-B                  & 28.84 & \multicolumn{1}{c|}{29.97} & \multicolumn{1}{c|}{-}     & \multicolumn{1}{c|}{31.62}       & 31.78       & \multicolumn{1}{c|}{31.61}          & \multicolumn{1}{c|}{{\ul 31.80}} & \textbf{31.94} \\ \hline
\multirow{12}{*}{BSD500} & \multirow{3}{*}{10} & PSNR                    & 25.92 & \multicolumn{1}{c|}{26.42} & \multicolumn{1}{c|}{27.61} & \multicolumn{1}{c|}{27.57}       & 27.62       & \multicolumn{1}{c|}{27.74}          & \multicolumn{1}{c|}{{\ul 27.85}} & \textbf{28.07} \\ \cline{3-11} 
                         &                     & SSIM                    & 0.739 & \multicolumn{1}{c|}{0.777} & \multicolumn{1}{c|}{0.796} & \multicolumn{1}{c|}{0.792}       & 0.789       & \multicolumn{1}{c|}{\textbf{0.802}} & \multicolumn{1}{c|}{{\ul 0.799}} & {\ul 0.799}    \\ \cline{3-11} 
                         &                     & PSNR-B                  & 24.22 & \multicolumn{1}{c|}{25.74} & \multicolumn{1}{c|}{27.22} & \multicolumn{1}{c|}{27.31}       & 27.34       & \multicolumn{1}{c|}{27.47}          & \multicolumn{1}{c|}{{\ul 27.52}} & \textbf{27.74} \\ \cline{2-11} 
                         & \multirow{3}{*}{20} & PSNR                    & 28.29 & \multicolumn{1}{c|}{28.84} & \multicolumn{1}{c|}{29.90} & \multicolumn{1}{c|}{29.97}       & 30.03       & \multicolumn{1}{c|}{30.01}          & \multicolumn{1}{c|}{{\ul 30.14}} & \textbf{30.36} \\ \cline{3-11} 
                         &                     & SSIM                    & 0.825 & \multicolumn{1}{c|}{0.856} & \multicolumn{1}{c|}{0.863} & \multicolumn{1}{c|}{0.861}       & 0.862       & \multicolumn{1}{c|}{\textbf{0.869}} & \multicolumn{1}{c|}{0.867}       & {\ul 0.868}    \\ \cline{3-11} 
                         &                     & PSNR-B                  & 26.45 & \multicolumn{1}{c|}{26.48} & \multicolumn{1}{c|}{29.20} & \multicolumn{1}{c|}{29.33}       & 29.51       & \multicolumn{1}{c|}{29.53}          & \multicolumn{1}{c|}{{\ul 29.56}} & \textbf{29.77} \\ \cline{2-11} 
                         & \multirow{3}{*}{30} & PSNR                    & 29.64 & \multicolumn{1}{c|}{30.12} & \multicolumn{1}{c|}{-}     & \multicolumn{1}{c|}{31.16}       & 31.25       & \multicolumn{1}{c|}{31.33}          & \multicolumn{1}{c|}{{\ul 31.45}} & \textbf{31.68} \\ \cline{3-11} 
                         &                     & SSIM                    & 0.863 & \multicolumn{1}{c|}{0.888} & \multicolumn{1}{c|}{-}     & \multicolumn{1}{c|}{0.885}       & 0.888       & \multicolumn{1}{c|}{\textbf{0.898}} & \multicolumn{1}{c|}{{\ul 0.897}} & \textbf{0.898} \\ \cline{3-11} 
                         &                     & PSNR-B                  & 27.78 & \multicolumn{1}{c|}{28.20} & \multicolumn{1}{c|}{-}     & \multicolumn{1}{c|}{30.74}       & {\ul 30.75} & \multicolumn{1}{c|}{30.70}          & \multicolumn{1}{c|}{30.72}       & \textbf{30.91} \\ \cline{2-11} 
                         & \multirow{3}{*}{40} & PSNR                    & 30.59 & \multicolumn{1}{c|}{30.96} & \multicolumn{1}{c|}{-}     & \multicolumn{1}{c|}{32.39}       & {\ul 32.44} & \multicolumn{1}{c|}{32.25}          & \multicolumn{1}{c|}{32.36}       & \textbf{32.58} \\ \cline{3-11} 
                         &                     & SSIM                    & 0.885 & \multicolumn{1}{c|}{0.905} & \multicolumn{1}{c|}{-}     & \multicolumn{1}{c|}{0.911}       & {\ul 0.913} & \multicolumn{1}{c|}{\textbf{0.915}} & \multicolumn{1}{c|}{0.913}       & {\ul 0.914}    \\ \cline{3-11} 
                         &                     & PSNR-B                  & 28.74 & \multicolumn{1}{c|}{29.35} & \multicolumn{1}{c|}{-}     & \multicolumn{1}{c|}{{\ul 31.56}} & 31.53       & \multicolumn{1}{c|}{31.5}           & \multicolumn{1}{c|}{31.52}       & \textbf{31.69} \\ \hline
\end{tabular}%
\end{table*}

\begin{table*}[t]
\renewcommand{\arraystretch}{1.25}
\caption{Quantitative comparisons on LIU4K \cite{liu2020comprehensive}. Please note that the multi-model methods train a specific model for each quality factor, while the single-model methods train one model for all quality factors.}
\centering
\label{tab:quantitative_comparison_liu}
\begin{tabular}{|c|c|c|c|cccc|ccc|}
\hline
\multirow{2}{*}{Dataset} & \multirow{2}{*}{QF} & \multirow{2}{*}{Metric} &       & \multicolumn{4}{c|}{Multi   Models}                                                                      & \multicolumn{3}{c|}{Single   Model}                                                     \\ \cline{4-11} 
                         &                     &                         & JPEG  & \multicolumn{1}{c|}{ARCNN \cite{ARCNNdong2015compression}} & \multicolumn{1}{c|}{DMCNN \cite{zhang2018dmcnn}} & \multicolumn{1}{c|}{IPT \cite{chen2021pre}}         & SWINIR \cite{liang2021swinir}      & \multicolumn{1}{c|}{QGAC \cite{ehrlich2020quantization}}           & \multicolumn{1}{c|}{FBCNN \cite{jiang2021towards}}       & Ours           \\ \hline
\multirow{12}{*}{LIU4K}  & \multirow{3}{*}{10} & PSNR                    & 29.49 & \multicolumn{1}{c|}{30.84} & \multicolumn{1}{c|}{31.76} & \multicolumn{1}{c|}{31.94}       & 31.97       & \multicolumn{1}{c|}{32.10}          & \multicolumn{1}{c|}{{\ul 32.16}} & \textbf{32.44} \\ \cline{3-11} 
                         &                     & SSIM                    & 0.809 & \multicolumn{1}{c|}{0.841} & \multicolumn{1}{c|}{0.859} & \multicolumn{1}{c|}{0.861}       & 0.860       & \multicolumn{1}{c|}{{\ul 0.865}}    & \multicolumn{1}{c|}{{\ul 0.865}}       & \textbf{0.867} \\ \cline{3-11} 
                         &                     & PSNR-B                  & 28.32 & \multicolumn{1}{c|}{30.54} & \multicolumn{1}{c|}{31.65} & \multicolumn{1}{c|}{31.81}       & 31.82       & \multicolumn{1}{c|}{32.04}          & \multicolumn{1}{c|}{{\ul 32.05}} & \textbf{32.32} \\ \cline{2-11} 
                         & \multirow{3}{*}{20} & PSNR                    & 32.42 & \multicolumn{1}{c|}{33.20} & \multicolumn{1}{c|}{34.13} & \multicolumn{1}{c|}{34.37}       & 34.40       & \multicolumn{1}{c|}{34.60}          & \multicolumn{1}{c|}{{\ul 34.73}} & \textbf{34.94} \\ \cline{3-11} 
                         &                     & SSIM                    & 0.866 & \multicolumn{1}{c|}{0.881} & \multicolumn{1}{c|}{0.895} & \multicolumn{1}{c|}{0.896}       & 0.895       & \multicolumn{1}{c|}{{\ul 0.900}}    & \multicolumn{1}{c|}{0.899} & \textbf{0.901} \\ \cline{3-11} 
                         &                     & PSNR-B                  & 31.15 & \multicolumn{1}{c|}{32.97} & \multicolumn{1}{c|}{33.90} & \multicolumn{1}{c|}{34.14}       & 34.14       & \multicolumn{1}{c|}{34.49}          & \multicolumn{1}{c|}{{\ul 34.54}} & \textbf{34.69} \\ \cline{2-11} 
                         & \multirow{3}{*}{30} & PSNR                    & 33.87 & \multicolumn{1}{c|}{34.26} & \multicolumn{1}{c|}{-}     & \multicolumn{1}{c|}{35.62}       & 35.63       & \multicolumn{1}{c|}{35.85}          & \multicolumn{1}{c|}{{\ul 36.00}} & \textbf{36.16} \\ \cline{3-11} 
                         &                     & SSIM                    & 0.892 & \multicolumn{1}{c|}{0.901} & \multicolumn{1}{c|}{-}     & \multicolumn{1}{c|}{0.914}       & 0.914       & \multicolumn{1}{c|}{{\ul 0.917}}          & \multicolumn{1}{c|}{{\ul 0.917}} & \textbf{0.919} \\ \cline{3-11} 
                         &                     & PSNR-B                  & 32.57 & \multicolumn{1}{c|}{34.07} & \multicolumn{1}{c|}{-}     & \multicolumn{1}{c|}{35.32}       & 35.29       & \multicolumn{1}{c|}{35.70}          & \multicolumn{1}{c|}{{\ul 35.75}} & \textbf{35.85} \\ \cline{2-11} 
                         & \multirow{3}{*}{40} & PSNR                    & 34.86 & \multicolumn{1}{c|}{34.94} & \multicolumn{1}{c|}{-}     & \multicolumn{1}{c|}{36.35}       & 36.40       & \multicolumn{1}{c|}{36.64}          & \multicolumn{1}{c|}{{\ul 36.83}} & \textbf{36.96} \\ \cline{3-11} 
                         &                     & SSIM                    & 0.907 & \multicolumn{1}{c|}{0.913} & \multicolumn{1}{c|}{-}     & \multicolumn{1}{c|}{0.924}       & 0.924       & \multicolumn{1}{c|}{0.927}          & \multicolumn{1}{c|}{{\ul 0.928}} & \textbf{0.929} \\ \cline{3-11} 
                         &                     & PSNR-B                  & 33.54 & \multicolumn{1}{c|}{34.76} & \multicolumn{1}{c|}{-}     & \multicolumn{1}{c|}{35.99}       & 36.00       & \multicolumn{1}{c|}{36.46}          & \multicolumn{1}{c|}{{\ul 36.51}} & \textbf{36.56} \\ \hline
\end{tabular}%
\end{table*}

\begin{table}[t!]
\renewcommand{\arraystretch}{1.25}
\caption{FLOPs, the number of parameters, and the running time on color images with $512 \times 512$ color images.}
\centering
\label{tab:efficiency_comparison}
\begin{tabular}{|c|c|c|c|}
\hline
Method & FLOPs (G) & \#Params (M) & Running Time (ms) \\ \hline
DMCNN \cite{zhang2018dmcnn} & 687.1     & 4.7    &  31.3  \\ \hline
IPT \cite{chen2021pre}   & 70243.5   & 114.2   & 5413.7  \\ \hline
SWINIR \cite{liang2021swinir}& 50768.5   & 11.5    & 3643.4  \\ \hline
QGAC \cite{ehrlich2020quantization}  & 249.9     & 0.0     & 113.1  \\ \hline
FBCNN \cite{jiang2021towards} & 728.9     & 71.9    & 13.0  \\ \hline
Ours   & 952.8     & 109.9   & 29.7  \\ \hline
\end{tabular}
\end{table}
\begin{figure}[t!]
    \centering
    \includegraphics[width=1.00\linewidth]{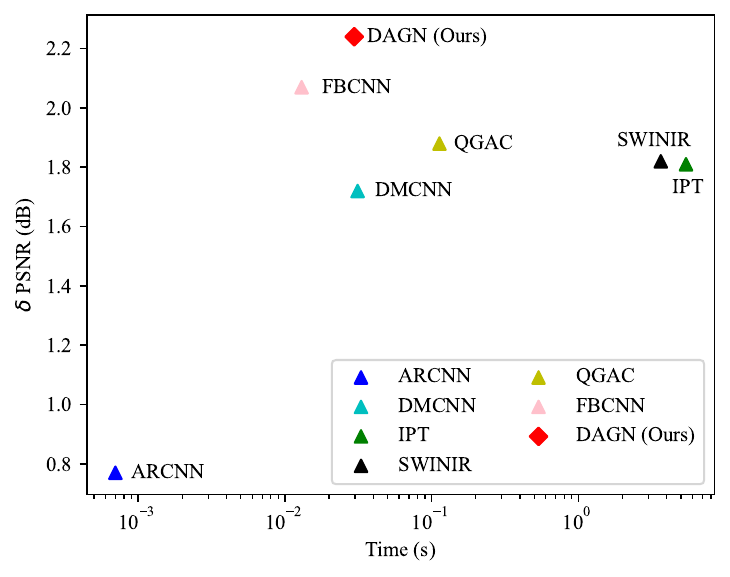}
    \caption{Average PSNR gains v.s the running times of different methods on the $200$ color images with size $512 \times 512$.}
    \label{fig:times}
\end{figure}
\begin{figure*}[t!]
    \centering
    \includegraphics[width=1.00\linewidth]{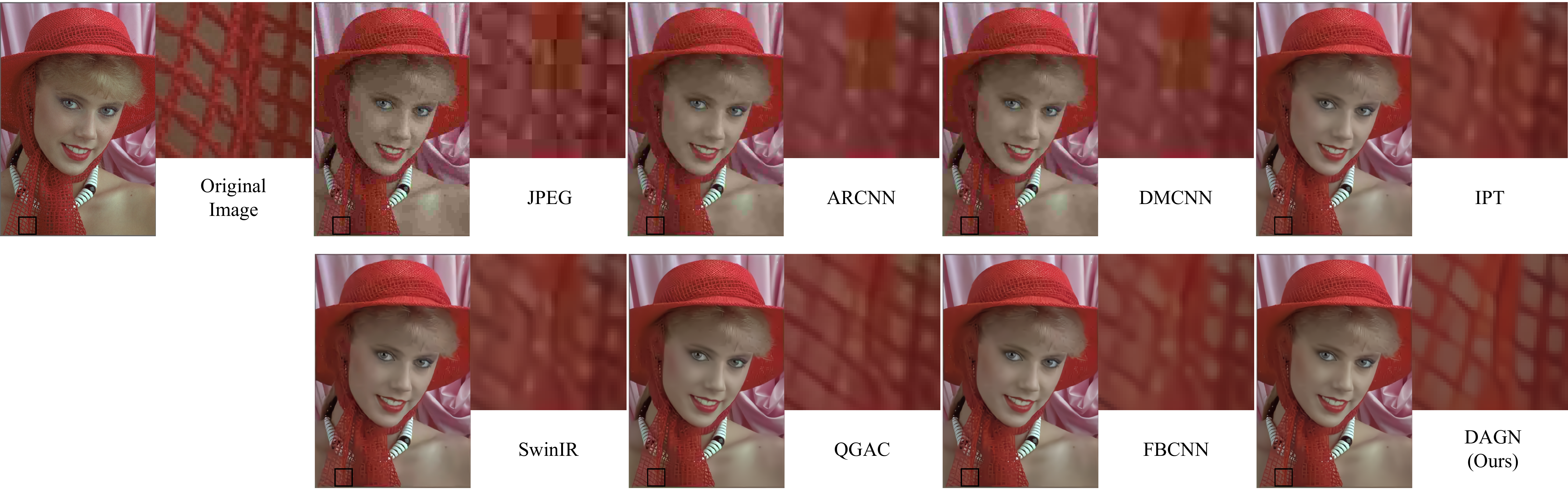}
    \caption{Qualitative comparisons on \emph{Womanhat} at $\mathrm{QF}=10$. For comparisons, we highlight the regions of interest. Original Image, PSNR $|$ SSIM $|$ PSNR-B. JPEG 27.68$|$ 0.722$|$ 26.45. ARCNN \cite{ARCNNdong2015compression} 28.36$|$ 0.753$|$ 28.34. DMCNN \cite{zhang2018dmcnn} 29.21$|$ 0.777$|$ 29.16. IPT \cite{chen2021pre} 29.36$|$ 0.778$|$ 29.32. SwinIR \cite{liang2021swinir} 29.47$|$ 0.780$|$ 29.42. QGAC \cite{ehrlich2020quantization} 29.78$|$ 0.791$|$ 29.76. FBCNN \cite{jiang2021towards} 29.76$|$ 0.786$|$ 29.74. DAGN (ours) 30.03$|$ 0.793$|$ 30.03.}
    \label{fig:vs_10}
\end{figure*}

\section{Experiments}
\label{sec:exp}
To demonstrate the superiority of DAGN, we conduct quantitative and qualitative evaluations on synthetic JPEG datasets and real-world cases. We compare the proposed DFGN with state-of-the-art methods, including the pioneer DNN-based method ARCNN \cite{ARCNNdong2015compression}, a dual-domain learning-based method DMCNN \cite{zhang2018dmcnn}, two transformer-based methods IPT \cite{chen2021pre} and SwinIR \cite{liang2021swinir}, a quantization table-based method QGAC \cite{ehrlich2020quantization}, and a blind method FBCNN \cite{jiang2021towards}. 

\subsection{Experimental Datasets and Setting}
\subsubsection{Datasets}
Following \cite{ehrlich2020quantization,jiang2021towards}, we employ DIV2K \cite{agustsson2017ntire} and Flickr2K \cite{radu2017ntire} as training datasets. The former consists of 900 images, and the latter contains 2,650 images.
For fair comparisons, the JPEG encoder in MATLAB is employed to compress images used in training and evaluation. Following prevailing methods, we conducted experiments on color JPEG images. 

\subsubsection{Training setting of the compression-insensitive encoder and compression-sensitive encoder}
During the compression-insensitive and sensitive encoder training, we randomly sample QF from 10 to 95 to get compressed images. 
{
The batch size is set to 32, and each image is resized to $256\times256$. The optimizer is the Adam \cite{kingma2014adam} with a $10^{-4}$ learning rate. The learning rate is set to $5\times10^{-4}$ and drops by $0.1$ at 5 and 10 epochs. The total epoch is set to 15.
}
\subsubsection{Training setting of DAGN}
We randomly extract a $96 \times 96$ patch for each training image and compress the patch with QF randomly sampled from 10 to 95.
{
We utilize Adam \cite{kingma2014adam} to optimize the parameters of DAGN by minimizing the $\mathcal{L}_{1}$ pixel loss. 
The batch size is set to 64. 
}
The learning rate is set to $1 \times 10^{-4}$ and decays by a factor of $0.5$ at $8 \times 10^{3}$ epochs. The total epochs are set to $1 \times 10^{4}$. 

\begin{figure*}[t!]
    \centering
    \includegraphics[width=1.00\linewidth]{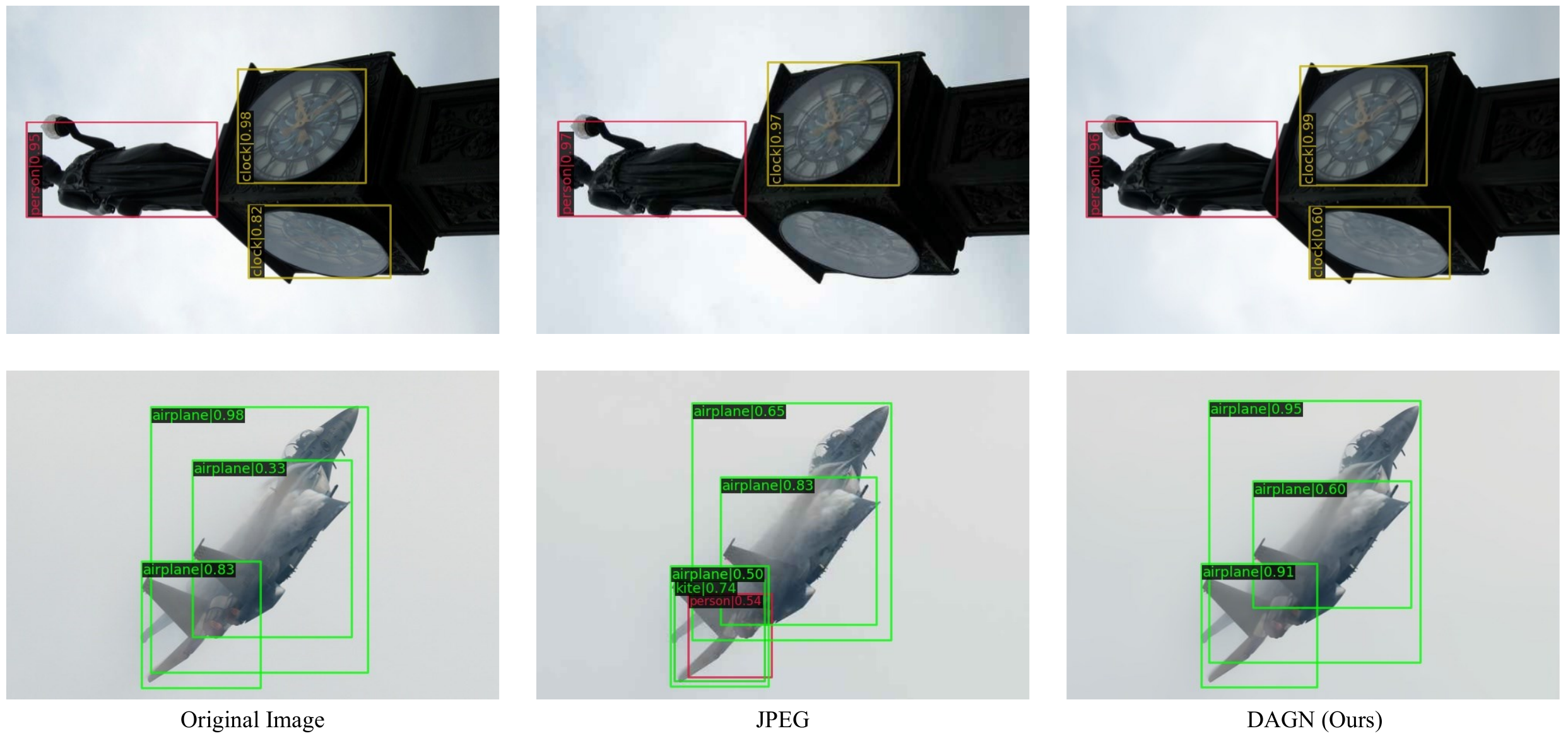}
    \caption{Object detection results on two samples in MS COCO \cite{Lin2014Microsoft}, including the original images + object detection \cite{Ren_2017_fasterrcnn}, JPEG images ($\mathrm{QF}=40$) + object detection \cite{Ren_2017_fasterrcnn}, and the compression artifacts results of DAGN + object detection \cite{Ren_2017_fasterrcnn}.}
    \label{fig:vs_od40}
\end{figure*}
\subsection{Comparisons with State-of-the-art Methods}
\begin{figure*}[t]
    \centering
    \includegraphics[width=1.00\linewidth]{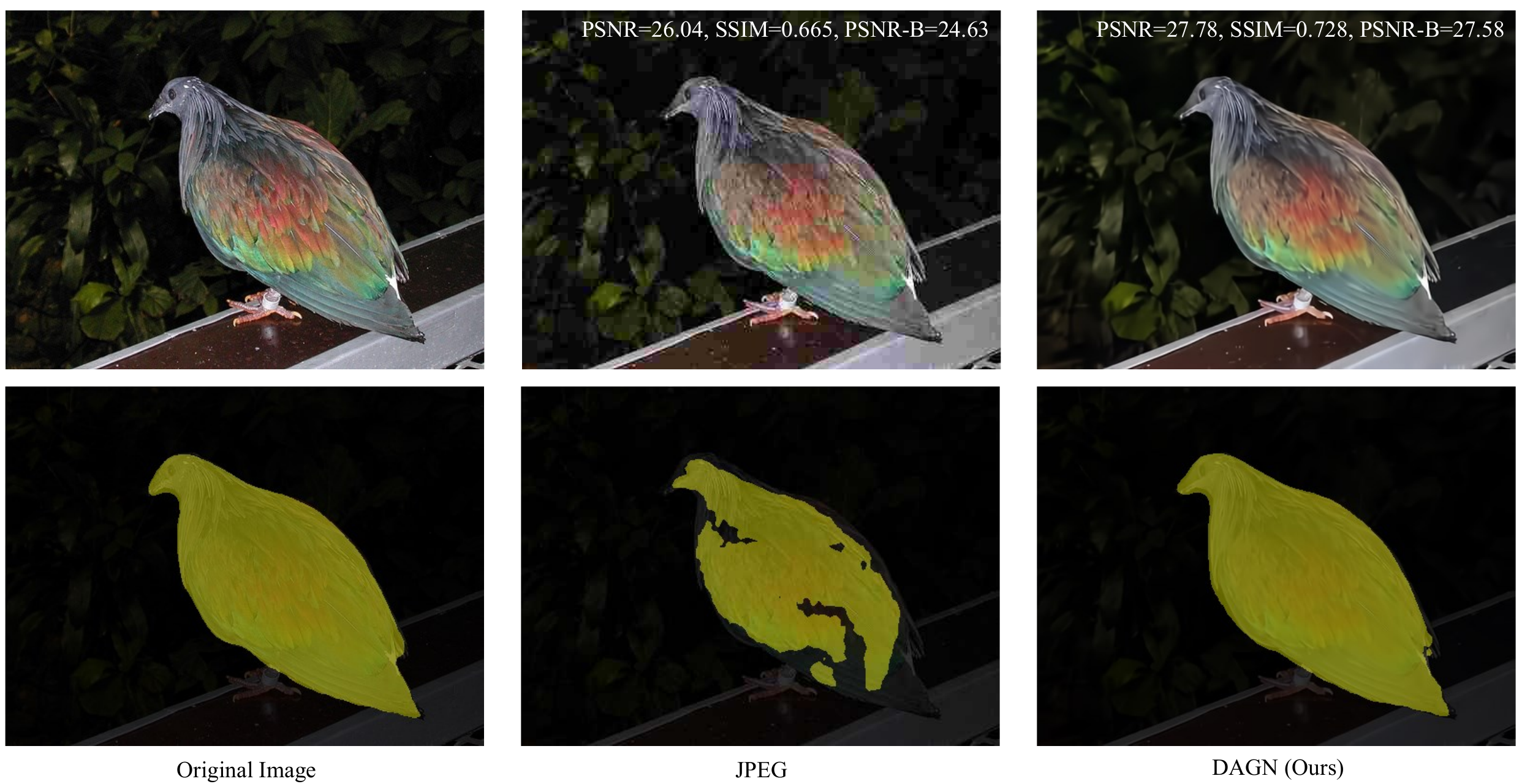}
    \caption{Comparisons of segmentation results on PASCAL VOC \cite{everingham2010pascal}. From left axis to right: the original images + semantic segmentation \cite{zhao2017pspnet}, JPEG images ($\mathrm{QF}=10$) + semantic segmentation \cite{zhao2017pspnet}, and the compression artifacts results of DAGN + semantic segmentation \cite{zhao2017pspnet}.}
    \label{fig:vs_ss10}
\end{figure*}

\subsubsection{Quantitative Comparisons}
We first report the quantitative assessment result of different methods on two widely used datasets, \emph{i.e.}, LIVE1 and BSD500. The peak signal-to-noise ratio (PSNR), structural similarity (SSIM) \cite{zhou2004ssim}, and peak signal-to-noise ratio including blocking effects (PSNR-B) \cite{yim2011psnrb}, are conducted for quantitative evaluations. Note that the PSNR-B is specifically designed and more sensitive to blocking artifacts (one of the compression artifacts) than PSNR and SSIM. 

~\tabref{tab:quantitative_comparison} reports the quantitative results. It is clear that DAGN achieves the best overall result, with significant improvements over other algorithms.
For instance, compared with the first deep neural network-based method ARCNN \cite{ARCNNdong2015compression}, DAGN achieves (1.52 dB, 0.037, and 1.38 dB) improvements in (PSNR, SSIM, and PSNR-B) on LIVE1 and (1.65 dB, 0.022, and 2.00 dB) on BSD500 at $\mathrm{QF}=10$. 
Under the same condition, DAGN outperforms the dual-domain learning-based method DMCNN \cite{zhang2018dmcnn} on LIVE1 and BSD500 by (0.77 dB, 0.005, 0.67 dB) and (0.46 dB, 0.003, 0.52 dB), respectively.
Compared with the transformer-based method SwinIR \cite{liang2021swinir}, DAGN obtains (0.50 dB, 0.011, 0.47 dB) gains on LIVE1 and (0.45 dB, 0.010, 0.40 dB) on BSD500 at $\mathrm{QF}=10$, respectively. It should be noted that ARCNN, DMCNN, and SwinIR train one specific network for each QF, while DAGN trains a network for all QFs.
Compared with the flexible artifacts reduction method FBCNN \cite{jiang2021towards}, which also train a network for all QFs, DAGN obtains (0.16 dB, 0.005, and 0.17 dB) gains in (PSNR, SSIM, and PSNR-B) on LIVE1 and (0.22 dB, 0.001, 0.21 dB) on BSD500 at $\mathrm{QF}=20$. Similar results could be found at other QFs. These results demonstrate the superiority of the proposed DAGN over state-of-the-art learning-based methods.
{
Besides, to verify the performance of our method on 4K images, we evaluate DAGN on LIU4K \cite{liu2020comprehensive}, which contains 1500 4K images for training and 80 4K images for validation. Table \ref{tab:quantitative_comparison_liu} shows the results, which also demonstrate the superiority of DAGN.
}

\subsubsection{Qualitative Comparisons}
Besides the quantitative assessment, we present qualitative comparisons.~\figref{fig:vs_10} shows a sample at $\mathrm{QF}=10$. For comparison, the regions of interest are highlighted. It could be seen that the compression artifacts obviously exist in JPEG images. Although these artifacts could be repressed to different degrees by compression artifacts reduction methods, restoring structures and details is a challenge for the competing methods.
By contrast, the proposed DAGN not only eliminates compression artifacts effectively but also reconstructs more pleasant details and textures than other competing methods.

\subsubsection{Efficiency Comparisons}

{
Table \ref{tab:efficiency_comparison} reports FLOPs, the number of parameters, and the running time of the above methods.
Specifically, each method is employed to process $200$ $512\times 512$ color images, and the experiments are conducted on a single NVIDIA Tesla P100-SXM2-16 GB GPU. 
}
Moreover, we report the running time and the gains of PSNR in~\figref{fig:times}. It could be seen that DAGN outperforms ARCNN by a great margin (over 1.4 dB) at the expense of running time. Compared with DMCNN, QGAC, and FBCNN, whose running times are similar to that of DAGN, it obtains 0.26 dB, 0.25 dB, and 0.19 dB improvements, respectively. Besides, DAGN impressively uses less running time than the transformer-based methods, \emph{i.e.}, IPT and SwinIR. 
{
It could be observed that our proposed method introduces affordable computation costs but surpasses the prevailing methods by a margin, which indicates that our proposed method achieves a good trade-off considering the performance and time costs.
}
 
\subsection{Evaluations on Computer Vision Tasks}
\begin{table*}[t]
\renewcommand{\arraystretch}{1.25}
\caption{Comparisons on object detection and semantic segmentation. Please note that the multi-model methods train a specific model for each quality factor, while the single-model methods train one model for all quality factors.}
\centering
\label{tab:task_comparison}
\begin{tabular}{|c|c|c|c|cccc|ccc|}
\hline
\multirow{2}{*}{Task}                                                            & \multirow{2}{*}{Metric}   & \multirow{2}{*}{QF} &      & \multicolumn{4}{c|}{Multi-model Methods}                                                                & \multicolumn{3}{c|}{Single-model Methods}                                                 \\ \cline{4-11} 
                                            &                     &                         & JPEG  & \multicolumn{1}{c|}{ARCNN \cite{ARCNNdong2015compression}} & \multicolumn{1}{c|}{DMCNN \cite{zhang2018dmcnn}} & \multicolumn{1}{c|}{IPT \cite{chen2021pre}}         & SWINIR \cite{liang2021swinir}      & \multicolumn{1}{c|}{QGAC \cite{ehrlich2020quantization}}           & \multicolumn{1}{c|}{FBCNN \cite{jiang2021towards}}       & Ours            \\ \hline
\multirow{4}{*}{\begin{tabular}[c]{@{}c@{}}Object\\ detection\end{tabular}}      & \multirow{4}{*}{Bbox mAP} & 10                  & 9.5  & \multicolumn{1}{c|}{19.3}  & \multicolumn{1}{c|}{22.1}  & \multicolumn{1}{c|}{23.7} & 23.9       & \multicolumn{1}{c|}{{\ul 24.7}} & \multicolumn{1}{c|}{24.5}       & \textbf{26.1} \\ \cline{3-11} 
                                                                                 &                           & 20                  & 23.8 & \multicolumn{1}{c|}{29.1}  & \multicolumn{1}{c|}{29.2}  & \multicolumn{1}{c|}{29.6} & 29.8       & \multicolumn{1}{c|}{29.8}       & \multicolumn{1}{c|}{{\ul 30.5}} & \textbf{31.6} \\ \cline{3-11} 
                                                                                 &                           & 30                  & 30.0 & \multicolumn{1}{c|}{31.7}  & \multicolumn{1}{c|}{31.9}  & \multicolumn{1}{c|}{32.1} & 32.2       & \multicolumn{1}{c|}{32.0}       & \multicolumn{1}{c|}{{\ul 32.7}} & \textbf{33.7} \\ \cline{3-11} 
                                                                                 &                           & 40                  & 32.2 & \multicolumn{1}{c|}{32.7}  & \multicolumn{1}{c|}{32.9}  & \multicolumn{1}{c|}{33.1} & 33.3       & \multicolumn{1}{c|}{{\ul 33.4}} & \multicolumn{1}{c|}{{\ul 33.4}} & \textbf{34.7} \\ \hline
\multirow{4}{*}{\begin{tabular}[c]{@{}c@{}}Semantic\\ segmentation\end{tabular}} & \multirow{4}{*}{mean IoU} & 10                  & 51.8 & \multicolumn{1}{c|}{55.9}  & \multicolumn{1}{c|}{56.2}  & \multicolumn{1}{c|}{56.7} & 56.5       & \multicolumn{1}{c|}{56.8}       & \multicolumn{1}{c|}{{\ul 56.9}} & \textbf{57.9} \\ \cline{3-11} 
                                                                                 &                           & 20                  & 66.9 & \multicolumn{1}{c|}{67.4}  & \multicolumn{1}{c|}{67.8}  & \multicolumn{1}{c|}{68.2} & {\ul 68.3} & \multicolumn{1}{c|}{68.2}       & \multicolumn{1}{c|}{{\ul 68.3}} & \textbf{69.3} \\ \cline{3-11} 
                                                                                 &                           & 30                  & 71.7 & \multicolumn{1}{c|}{72.3}  & \multicolumn{1}{c|}{72.0}  & \multicolumn{1}{c|}{72.1} & {\ul 72.5} & \multicolumn{1}{c|}{{\ul 72.5}} & \multicolumn{1}{c|}{72.3}       & \textbf{73.1} \\ \cline{3-11} 
                                                                                 &                           & 40                  & 72.9 & \multicolumn{1}{c|}{73.1}  & \multicolumn{1}{c|}{72.8}  & \multicolumn{1}{c|}{73.2} & 73.2       & \multicolumn{1}{c|}{73.3}       & \multicolumn{1}{c|}{{\ul 73.7}} & \textbf{74.3} \\ \hline
\end{tabular}%
\end{table*}

{
Compression artifacts affect the performance of computer vision tasks apart from decreasing visual quality \cite{liu2020comprehensive, gao2021digital, chen2021feature}. In order to reduce the impact of compression artifacts, it may be helpful to suppress them during the preprocessing phase. As a result, we will examine the effectiveness of the proposed DAGN in this subsection, specifically regarding two common computer vision tasks, \emph{i.e.}, object detection, and semantic segmentation.
}
\subsubsection{Object Detection} 
The quantitative results are shown in~\tabref{tab:task_comparison}. We could observe that DAGN earns the best overall result, with significant improvements over the baseline JPEG.~\figref{fig:vs_od40} reports the results by Ren \emph{et al.}  \cite{Ren_2017_fasterrcnn} on original images, JPEG images, and the results of DAGN, respectively. It could be observed that compression artifacts could cause both missing and false detection, and the proposed DAGN could help boost the performance. For example, affected by compression artifacts, one clock is missed in the first axis of~\figref{fig:vs_od40}, and the airplane is wrongly recognized as a person in the second axis of~\figref{fig:vs_od40}. In contrast, 
with the help of DAGN, the detection results are nearly the same as the ones on the original image. 
\begin{figure*}[t]
    \centering
    \includegraphics[width=0.98\linewidth]{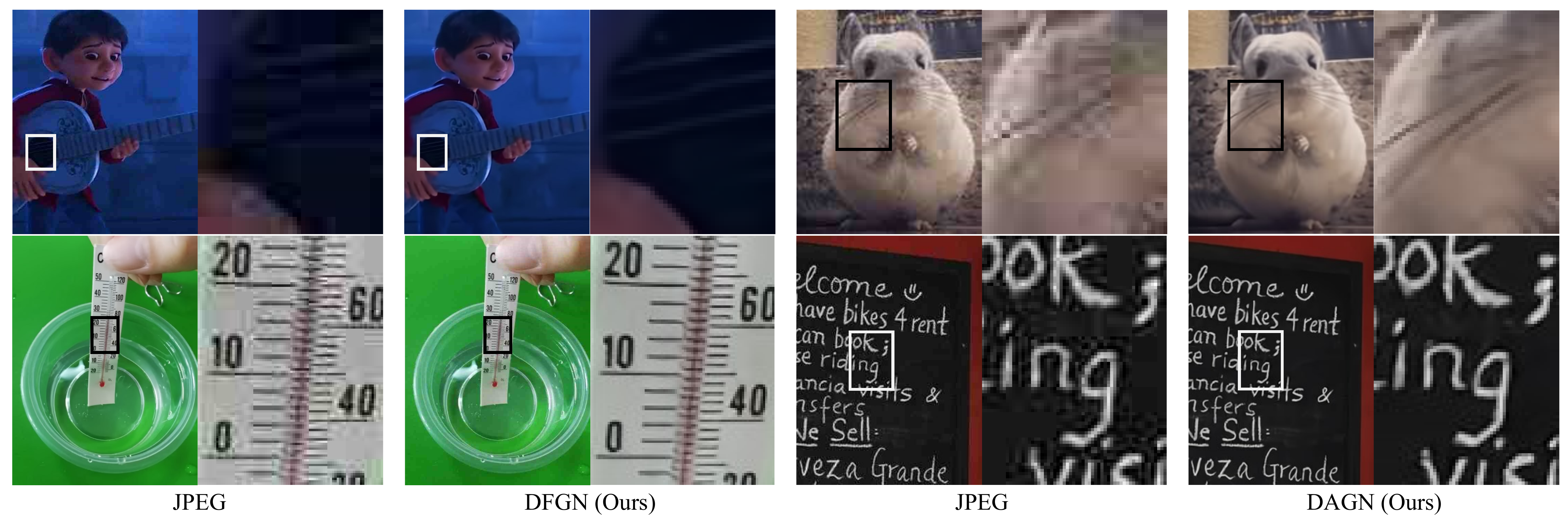}
    \caption{Comparison before and after applying DAGN on real-world cases. For comparison, the regions of interest are highlighted. It could be observed that DAGN could restore patterns with better visual quality.}
    \label{fig:vs_rw}
\end{figure*}
\begin{table*}[tb!]
\renewcommand{\arraystretch}{1.25}
\caption{Quantitative comparison with different cases (\emph{i.e.,}): Baseline: Only base encoder and guided decoder; Baseline w. CIGM: Base encoder, guided decoder, and compression-insensitive guidance module; Baseline w. CSGM: base encoder, guided decoder, and compression-sensitive guidance module; Baseline w. CFM: base encoder, guided decoder, and cross-feature fusion module. The best scores are \textbf{highlighted}.}
\centering
\label{tab:abs}
\begin{tabular}{|c|c|c|c|c|c|c|c|}
\hline
Dataset                  & QF                  & Metric & Baseline  &Baseline \emph{w.} CIGM   & Baseline \emph{w.} CSGM  & Baseline \emph{w.} CFM   & DAGN       \\ \hline
\multirow{12}{*}{LIVE1}  & \multirow{3}{*}{10} & PSNR   & 27.46 & 27.82  & 27.74 & 27.72 & \textbf{27.95} \\ \cline{3-8} 
                         &                     & SSIM   & 0.798 & 0.803  & 0.801 & 0.799 & \textbf{0.807} \\ \cline{3-8} 
                         &                     & PSNR-B & 27.11 & 27.57  & 27.50 & 27.47 & \textbf{27.70} \\ \cline{2-8} 
                         & \multirow{3}{*}{20} & PSNR   & 29.82 & 30.15  & 30.08 & 30.03 & \textbf{30.27} \\ \cline{3-8} 
                         &                     & SSIM   & 0.866 & 0.870  & 0.868 & 0.866 & \textbf{0.873} \\ \cline{3-8} 
                         &                     & PSNR-B & 29.34 & 29.73  & 29.68 & 29.6  & \textbf{29.87} \\ \cline{2-8} 
                         & \multirow{3}{*}{30} & PSNR   & 31.12 & 31.47  & 31.40 & 31.53 & \textbf{31.60} \\ \cline{3-8} 
                         &                     & SSIM   & 0.894 & 0.898  & 0.897 & 0.895 & \textbf{0.900} \\ \cline{3-8} 
                         &                     & PSNR-B & 30.57 & 30.95  & 30.92 & 30.83 & \textbf{31.10} \\ \cline{2-8} 
                         & \multirow{3}{*}{40} & PSNR   & 32.01 & 32.37  & 32.31 & 32.20 & \textbf{32.50} \\ \cline{3-8} 
                         &                     & SSIM   & 0.910 & 0.914  & 0.913 & 0.911 & \textbf{0.916} \\ \cline{3-8} 
                         &                     & PSNR-B & 31.44 & 31.81  & 31.77 & 31.64 & \textbf{31.94} \\ \hline
\multirow{12}{*}{BSD500} & \multirow{3}{*}{10} & PSNR   & 27.68 & 27.96  & 27.92 & 27.86 & \textbf{28.07} \\ \cline{3-8} 
                         &                     & SSIM   & 0.792 & 0.795 & 0.793 & 0.792 & \textbf{0.799} \\ \cline{3-8} 
                         &                     & PSNR-B & 27.26 & 27.64  & 27.6  & 27.54 & \textbf{27.74} \\ \cline{2-8} 
                         & \multirow{3}{*}{20} & PSNR   & 30.00 & 30.25  & 30.21 & 30.13 & \textbf{30.36} \\ \cline{3-8} 
                         &                     & SSIM   & 0.862 & 0.865  & 0.863 & 0.861 & \textbf{0.868} \\ \cline{3-8} 
                         &                     & PSNR-B & 29.36 & 29.67  & 29.65 & 29.54 & \textbf{29.77} \\ \cline{2-8} 
                         & \multirow{3}{*}{30} & PSNR   & 31.31 & 31.56  & 31.52 & 31.39 & \textbf{31.68} \\ \cline{3-8} 
                         &                     & SSIM   & 0.893 & 0.895  & 0.894 & 0.891 & \textbf{0.898} \\ \cline{3-8} 
                         &                     & PSNR-B & 30.54 & 30.82  & 30.81 & 30.65 & \textbf{30.91} \\ \cline{2-8} 
                         & \multirow{3}{*}{40} & PSNR   & 32.05 & 32.47  & 32.44 & 32.22 & \textbf{32.58} \\ \cline{3-8} 
                         &                     & SSIM   & 0.907 & 0.912  & 0.911 & 0.910 & \textbf{0.914} \\ \cline{3-8} 
                         &                     & PSNR-B & 31.20 & 31.61  & 31.60 & 31.36 & \textbf{31.69} \\ \hline
\end{tabular}
\end{table*}
\subsubsection{Semantic Segmentation}
The quantitative results are shown in~\tabref{tab:task_comparison}. It could be observed that DAGN earns the best result.
The semantic segmentation results by Zhao \emph{et al.} \cite{zhao2017pspnet} on original images, JPEG images, and the results of DAGN are reported in~\figref{fig:vs_ss10}. One could observe that compression artifacts decrease the visual quality and lead to poor segmentation. Then, we employ the proposed DAGN to mitigate the effect of compression artifacts. It could be seen that the semantic segmentation results from the compression artifacts reduction result produced by DAGN are close to those of the original image. This evidence verifies that our proposed DAGN significantly retains the high-level features for various downstream parsing tasks.

\subsection{Real-World Use Cases}
Besides the synthetic datasets, we evaluate the proposed DAGN in real-world use cases. Actually, most images shared on the Internet are compressed to save storage space. We collect 100 compressed images from the Internet and apply the proposed DAGN to reduce compression artifacts. 
~\figref{fig:vs_rw} shows four examples. Since there are no original images, the quantitative results are not shown. From~\figref{fig:vs_rw}, one could observe that there are several artifacts in the online compressed images, degrading the visual experience dramatically. The proposed DAGN not only significantly represses compression artifacts but also enacts pleasingly on restoring details.

\subsection{Ablation Study}
\subsubsection{The contribution of CIGM, SIGM, and CFM}
To further investigate the proposed DAGN, we conduct an ablation study to measure the contribution of the three components, including the compression-insensitive guidance module, compression-sensitive guidance module, and cross-feature fusion module.~\tabref{tab:abs} shows the comparisons between models with four different cases:
\begin{itemize}
    \item Baseline: Only the base encoder and guided decoder. Specifically, the guided decoder only takes baseline features as input, and $(\boldsymbol{\beta}_{i}, \boldsymbol{\gamma}_{i}), (\boldsymbol{\eta}_{i}, \boldsymbol{\epsilon}_{i}), i=1,2,3$ in the guided decoder are set to $(\boldsymbol{O}, \boldsymbol{J})$ and $(\boldsymbol{O}, \boldsymbol{J})$, where $\boldsymbol{O}$ and $\boldsymbol{J}$ indicate zero matrix and all-ones matrix, respectively.
    \item Baseline \emph{w.} CIGM: The base encoder, the guided decoder, and the compression-insensitive guidance module. Specifically, the guided decoder only takes baseline features as input, and $(\boldsymbol{\beta}_{i}, \boldsymbol{\gamma}_{i}), i=1,2,3$ are generated by the compression-insensitive guider from the compression-insensitive features, while $(\boldsymbol{\eta}_{i}, \boldsymbol{\epsilon}_{i}), i=1,2,3$ in the guided decoder are set to $(\boldsymbol{O}, \boldsymbol{J})$.
    \item Baseline \emph{w.} CSGM: The base encoder, the guided decoder, and the compression-sensitive guidance module. Specifically, the guided decoder only takes the baseline features as input, and $(\boldsymbol{\eta}_{i}, \boldsymbol{\epsilon}_{i}), i=1,2,3$ are generated by the compression-sensitive guider from the compression-sensitive features, while $(\boldsymbol{\beta}_{i}, \boldsymbol{\gamma}_{i}), i=1,2,3$ in the guided decoder are set to $(\boldsymbol{O}, \boldsymbol{J})$.
    \item Baseline \emph{w.} CFM: The base encoder, the guided decoder, and the cross-feature fusion module. Specifically, $(\boldsymbol{\beta}_{i}, \boldsymbol{\gamma}_{i}), (\boldsymbol{\eta}_{i}, \boldsymbol{\epsilon}_{i}), i=1,2,3$ in the guided decoder are set to $(\boldsymbol{O}, \boldsymbol{J})$ and $(\boldsymbol{O}, \boldsymbol{J})$, respectively.
\end{itemize}

It could be observed that the artifacts reduction performance has an obvious improvement with the three modules. 
{
For example, without the three modules, the baseline model achieves 27.46 dB, 0.798, 27.11 dB on LIVE1 and 27.68 dB, 0.792, 27.26 dB on BSD500 at $\mathrm{QF}=10$ in (PSNR, SSIM, PSNR-B). 
Firstly, CIGM considers compression-insensitive features and learns parameters for transforming feature representations in the decoding process. With the guidance of CIGM, the model could improve (0.36 dB, 0.005, 0.46 dB) on LIVE1 and (0.28 dB, 0.003, 0.38 dB) on BSD500.
Secondly, when adding the CSGM to the base encoder and guided decoder, the decoding process is adjusted by the parameters learned from compression-sensitive features. With the help of compression-sensitive features that reflect the degree of compression, the artifacts reduction performance is improved from (27.46 dB, 0.798, 27.11 dB) up to (27.74 dB, 0.801, 27.50 dB) on LIVE1 at $\mathrm{QF}=10$. 
Thirdly, we fuse compression-insensitive features with baseline features to incorporate the multi-scale semantic information from the compression-insensitive encoders. It could be observed that CFM could improve the artifacts reduction performance from (27.46 dB, 0.798, 27.11 dB) up to (27.72 dB, 0.799, 27.47 dB) on LIVE1 at $\mathrm{QF}=10$. 
Finally, combining the three modules further improves the artifacts reduction performance. 
}
These comparisons demonstrate the effectiveness of the compression-insensitive guidance module, compression-sensitive guidance module, and cross-feature fusion module.

\subsubsection{The influence of \texorpdfstring{$\lambda_{ci}$}{lambda ci} and \texorpdfstring{$\lambda_{cs}$}{lambda cs}}
\begin{figure}
    \centering
    \subfloat[]{\includegraphics[width=0.22\textwidth]{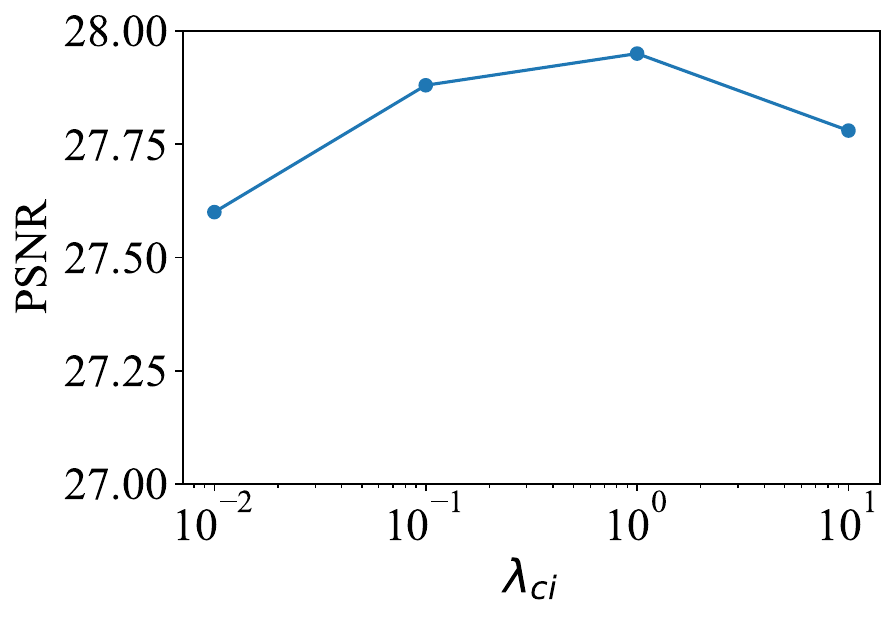}}
    \subfloat[]{\includegraphics[width=0.22\textwidth]{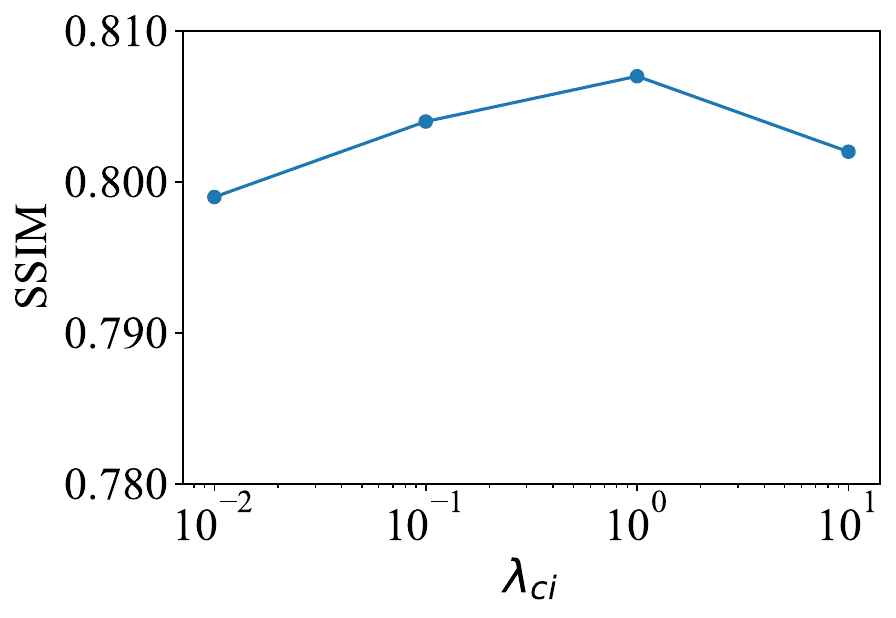}}
    \hfill
    \subfloat[]{\includegraphics[width=0.22\textwidth]{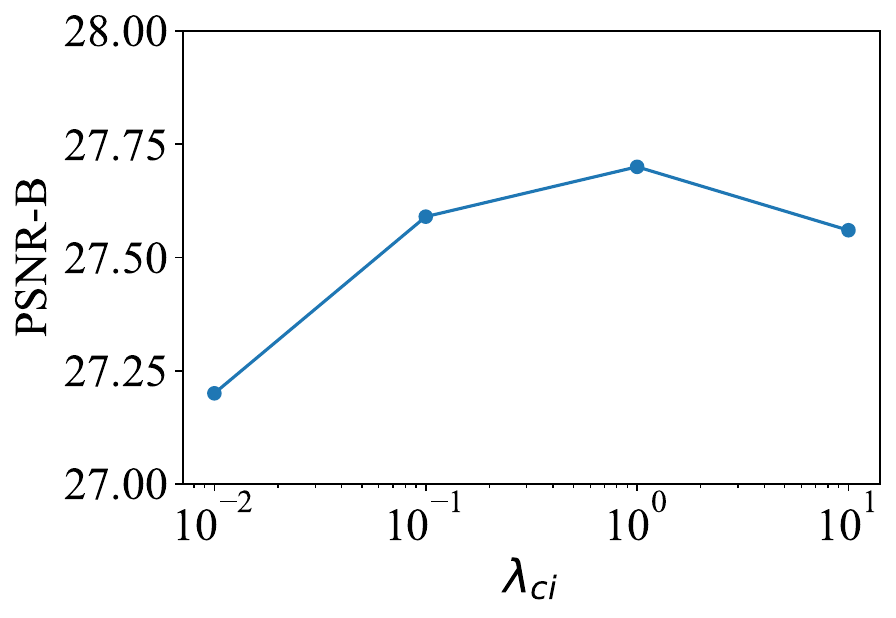}}
    \subfloat[]{\includegraphics[width=0.22\textwidth]{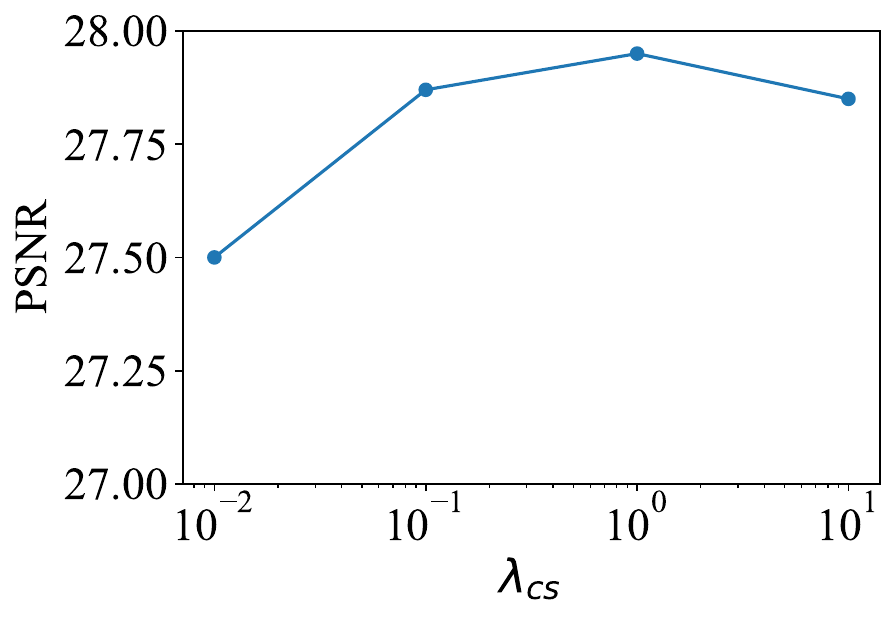}}
    \hfill
    \subfloat[]{\includegraphics[width=0.22\textwidth]{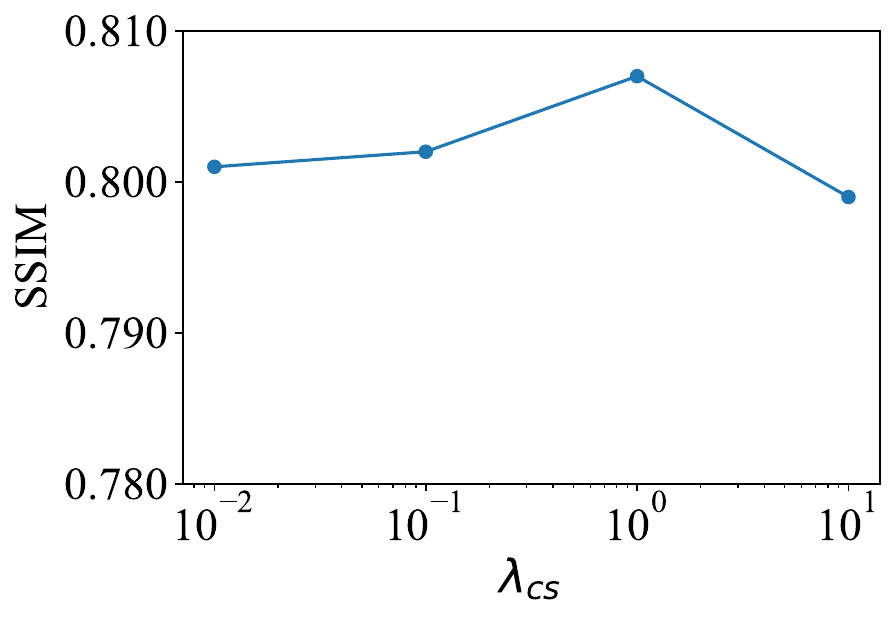}}
    \subfloat[]{\includegraphics[width=0.22\textwidth]{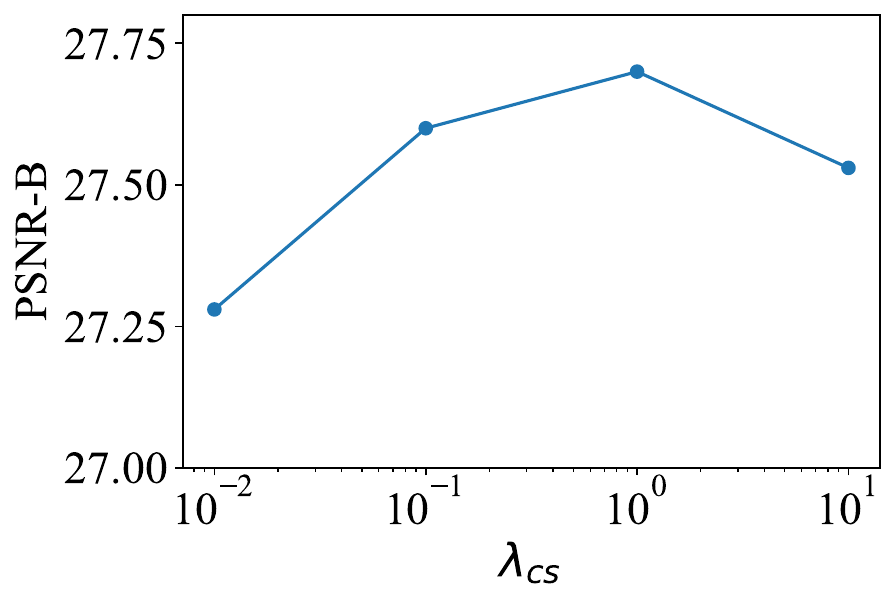}}
    \caption{The performance of DAGNs with different $\lambda_{ci}$ and $\lambda_{cs}$}
    \label{fig:cics}
\end{figure}

To explore the impact of $\lambda_{ci}$ and $\lambda_{cs}$, we train DAGNs in the case of $\lambda_{ci}$ and  $\lambda_{cs}$ with different values. ~\figref{fig:cics} shows the results. 
For $\lambda_{ci}$, we notice that DAGN achieves the best balance when $\lambda_{ci}=1$, and the model is not very sensitive to $\lambda_{ci}$. However, it is essential to avoid using values that are too large or too small since that could decrease the performance of DAGN. This occurs because the compression sensitive encoder focuses on the content-related features and ignores the compression-insensitive features when $\lambda_{ci}<1$. For $\lambda_{cs}$, similar conclusions could be drawn from~\figref{fig:cics}.

\section{Conclusion}
In this paper, {we start from a new view of intrinsic feature decoupling to solve the compression artifacts reduction problem. Toward this end, we}
propose a dual awareness guidance network for exploiting the intrinsic attributes to guide the learning of artifacts reduction. We first decouple the intrinsic attributes into compression-insensitive features and compression-sensitive features. Then compression-insensitive guidance module, compression-sensitive module, and cross-feature fusion module are designed to employ these features to reduce compression artifacts. Extensive experiments on synthetic and real compression datasets demonstrate the effectiveness and superiority of the proposed method. 
{
One future direction for this work is to explore decoupling learning to other image restoration tasks such as image denoising, super-resolution, and deblurring.
}

\label{sec:con}

\ifCLASSOPTIONcompsoc
  \section*{Acknowledgments}
\else
  \section*{Acknowledgment}
\fi
This work was supported in part by Key-Area Research and Development Program of Guangdong Province under Contract No. 2021B0101400002, and in part by the National Natural Science Foundation of China under Contract No. 62088102, and Contract No. 62202010.

\bibliographystyle{IEEEtran}
\bibliography{tip2022.bib}
\vspace{-10 mm}
\begin{IEEEbiography}[{\includegraphics[width=0.8in,height=1in,clip,keepaspectratio]{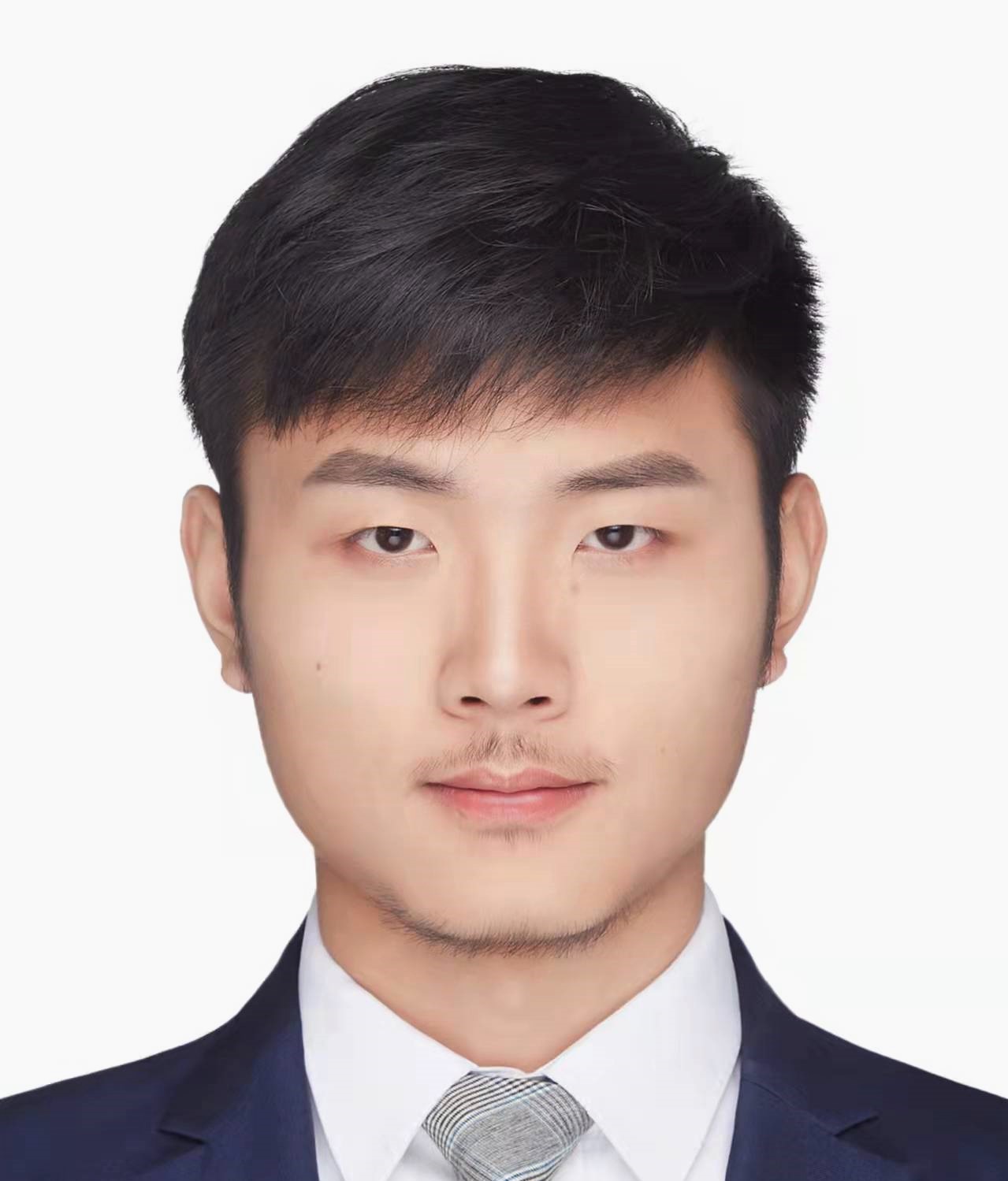}}]{Li Ma} received the B.S. degree from School of Mathematical Sciences and the Ph.D. degree from School of Computer Sciences, Peking University, Beijing, China, in 2016 and 2023. He joined Huawei Technologies Company, Ltd. in 2023. His research interests include video coding, video generation, and image/video understanding.
\end{IEEEbiography}
\vspace{-10 mm}
\begin{IEEEbiography}[{\includegraphics[width=0.8in,height=1in,clip,keepaspectratio]{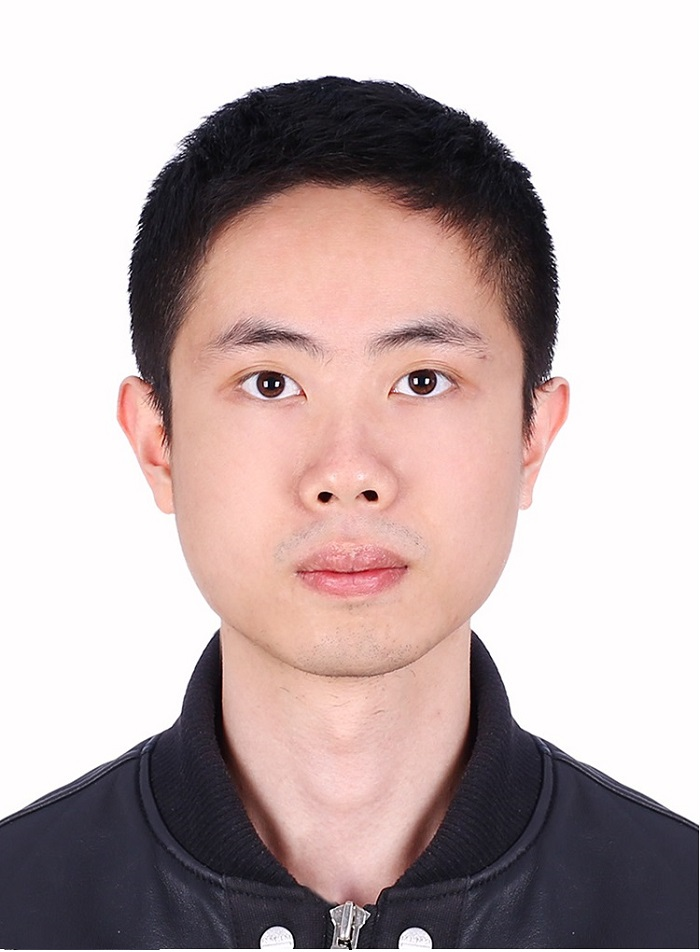}}]{Yifan Zhao} (Member, IEEE) is currently an Associated Professor with the School of Computer Science and Engineering, Beihang University, Beijing, China. He worked as a Boya Postdoc researcher with the School of Computer Science, Peking University. He received the B.E. degree from the Harbin Institute of Technology in Jul. 2016 and the Ph.D. degree from the School of Computer Science and Engineering, Beihang University, in Oct. 2021. His research interests include computer vision, VR/AR, and image/video understanding.
\end{IEEEbiography}
\vspace{-10mm}
\begin{IEEEbiography}[{\includegraphics[width=0.8in,height=1in,clip,keepaspectratio]{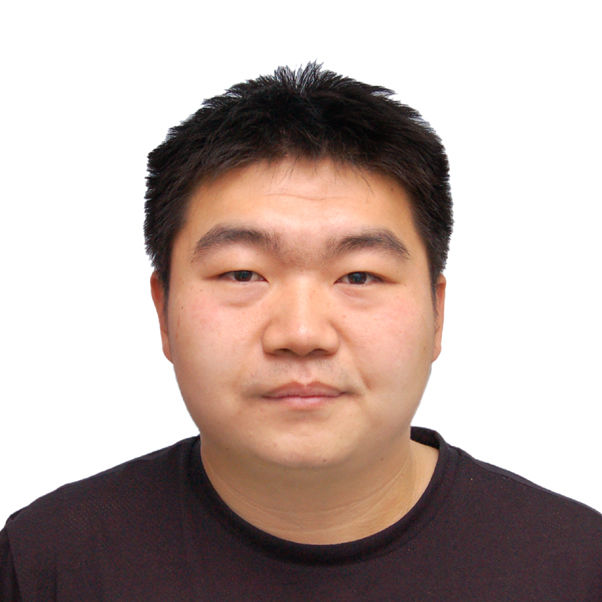}}]{Peixi Peng} received the PhD degree from  Peking University, in 2017, Beijing, China. He is currently a associate researcher with the School of Computer Science, Peking University, Beijing, China, and is also the assistant researcher of Artificial Intelligence Research Center, Peng Cheng Laboratory, Shenzhen, China.  He is the author or co-author of more than 20 technical articles in refereed journals such as IEEE
TPAMI, PR and conferences such as CVPR/ECCV/IJCAI/ACMMM/AAAI.
His research interests include computer vision, multimedia big data, and reinforcement learning.
\end{IEEEbiography}
\vspace{-10 mm}
\begin{IEEEbiography}[{\includegraphics[width=1.in,height=1in,clip,keepaspectratio]{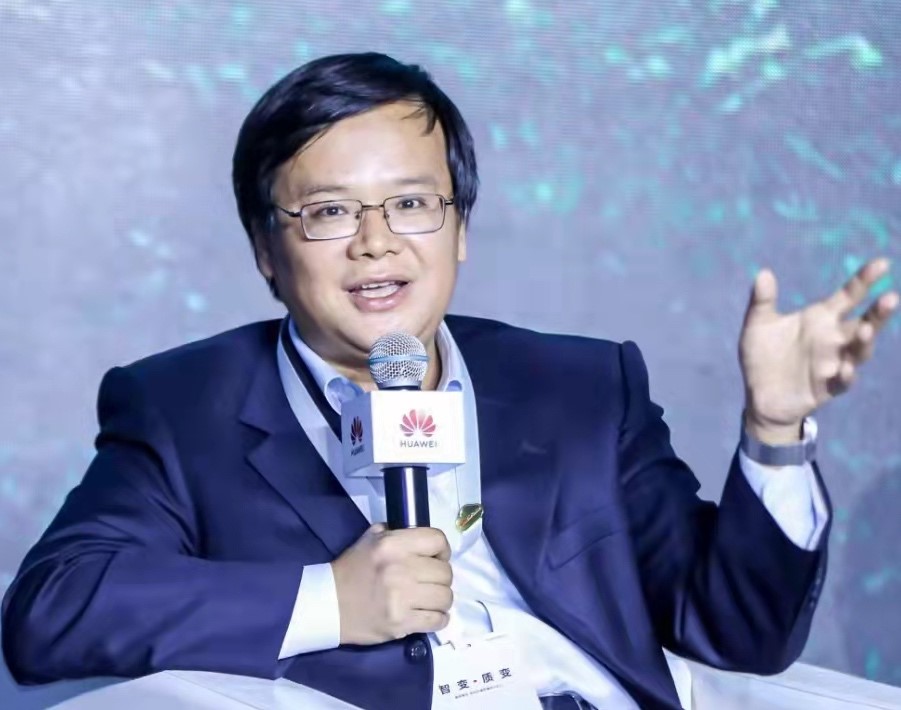}}]{Yonghong Tian} (S’00-M’06-SM’10-F’22) is currently the Dean of School of Electronics and Computer Engineering, a Boya Distinguished Professor with the School of Computer Science, Peking University, China, and is also the deputy director of Artificial Intelligence Research, PengCheng Laboratory, Shenzhen, China. His research interests include neuromorphic vision, distributed machine learning and AI for Science. He is the author or coauthor of over 350 technical articles in refereed journals and conferences. Prof. Tian was/is an Associate Editor of IEEE TCSVT (2018.1-2021.12), IEEE TMM (2014.8-2018.8), IEEE Multimedia Mag. (2018.1-2022.8), and IEEE Access (2017.1-2021.12). He co-initiated IEEE Int’l Conf. on Multimedia Big Data (BigMM) and served as the TPC Co-chair of BigMM 2015, and aslo served as the Technical Program Co-chair of IEEE ICME 2015, IEEE ISM 2015 and IEEE MIPR 2018/2019, and General Co-chair of IEEE MIPR 2020 and ICME2021. He is a TPC Member of more than ten conferences such as CVPR, ICCV, ACM KDD, AAAI, ACM MM and ECCV. He was the recipient of the Chinese National Science Foundation for Distinguished Young Scholars in 2018, two National Science and Technology Awards and three ministerial-level awards in China, and obtained the 2015 EURASIP Best Paper Award for Journal on Image and Video Processing, and the best paper award of IEEE BigMM 2018, and the 2022 IEEE SA Standards Medallion and SA Emerging Technology Award. He is a Fellow of IEEE, a senior member of CIE and CCF, a member of ACM.
\end{IEEEbiography}

\vfill
\end{document}